\documentclass[journal]{IEEEtran}

\usepackage{amssymb}
\usepackage{amsmath,bm}
\usepackage{algorithm}
\usepackage{algorithmic}
\usepackage{amsmath}
\usepackage{amssymb}
\usepackage{graphicx}
\usepackage{cite}
\usepackage{stfloats}

\newcommand{\PCR}{\mathrm{PCR6}}
\newcommand{\BetP}{\mathrm{BetP}}

\newcommand{\LNS}{\text{LNS-CR}}
\newcommand{\LNSA}{\text{LNSa-CR}}
\newcommand{\conj}{\mathrm{conj}}
\newcommand{\disj}{\mathrm{disj}}
\newcommand{\DP}{\mathrm{DP}}
\newcommand{\DS}{\mathrm{Dempster}}
\newcommand{\Ave}{\mathrm{Ave}}
\newcommand{\caut}{\mathrm{Cautious}}
\newcommand{\conf}{\mathrm{Conf}}
\newcommand{\x}{\bm{x}}

\newcommand{\ocap}{\operatornamewithlimits{\text{\textcircled{\scalebox{1.4}{\tiny{$\cap$}}}}}}
\newcommand{\ocup}{\operatornamewithlimits{\text{\textcircled{{\tiny{$\cup$}}}}}}
\newcounter{ExpNum}
\usecounter{ExpNum}
\newcommand{\Exp}{\noindent\textbf{Experiment~\arabic{ExpNum}}\refstepcounter{ExpNum}}
\usepackage[numbers]{natbib}
\usepackage{xcolor}
\newcommand{\change}{}

\begin{document}

\title{A belief combination rule for a large number of sources}

\author{\IEEEauthorblockN{Kuang Zhou$^\text{a}$,
Arnaud Martin$^\text{b}$,  and
Quan Pan$^\text{a}$\\
}
\IEEEauthorblockA{a. Northwestern Polytechnical University,
Xi'an, Shaanxi 710072, PR China. }\\
\IEEEauthorblockA{b. DRUID, IRISA, University of Rennes 1, Rue E. Branly, 22300 Lannion, France}
}

\maketitle

\begin{abstract}
The theory of belief functions is widely used for {data from multiple sources}.
Different evidence combination rules have been proposed in this framework
according to the properties of the sources to combine. However, most of
these combination rules are not efficient when there are  a large number of sources. This is due to either
the complexity  or the  existence of an absorbing
{element such as the total conflict mass function for the conjunctive based rules when applied on unreliable evidence}.
In this paper, based on the assumption that the majority of sources are reliable,
a combination rule for a large number of sources is proposed using a simple idea: the more common ideas the sources share, the more
reliable these sources are supposed to be. 
This rule is adaptable for aggregating  a large number of sources which may not  all be reliable. It will keep the spirit of the conjunctive rule to
reinforce the belief on the focal elements with
which the sources are in agreement. The mass on the empty set will  be kept as an indicator of the conflict.

The proposed rule, called {$\LNS$ (Conjunctive combination Rule for a Large Number of Sources)}, is evaluated on synthetic mass functions. The experimental results  verify
that  the rule  can be effectively used to combine a large number of mass functions and to
elicit the major opinion.
\end{abstract}

\begin{IEEEkeywords}
Theory of belief functions, big data, combination, large number of sources, reliability
\end{IEEEkeywords}

\IEEEpeerreviewmaketitle

\renewcommand{\thefootnote}{}
\section{Introduction}
\IEEEPARstart{I}n
{recent years,  Dempster--Shafer Theory (DST), also called the theory of belief functions, has gained
increasing attention in the scientific community as it allows to the deal with  the imprecise and uncertain information.
It has been applied in various domains\footnote{This paper is an extension and revision of \cite{zhou2017evidence}.},  such as data classification \cite{denoeux1995k,deng2016improved}, data
clustering \cite{masson2008ecm,zhou2016ecmdd},
social network analysis \cite{zhou2015median}, etc. }In complex environment, multiple stake-holders attempt to
reach a decision by combining several sources of information and aggregating their points of view by stressing common {agreement}.
{The theory of belief functions, which has provided many rules to combine information represented by mass functions \cite{Smets07a}, are widely used for decision making.
In real applications, 
there are usually a large number of sources.
Most of the existing combination rules are not {applicable} in this case, and cannot be
used to find the major opinion from many participants.}
\renewcommand{\thefootnote}{\fnsymbol{footnote}}
\setcounter{footnote}{0}

{One of the most famous combination rule in belief function framework is the Dempster's rule~\cite{Smets07a}.
{\citet{smets1990combination} proposed} a
modification of Dempster's rule,  often called ``conjunctive rule", where the empty set can be assigned with
a non-null mass under the Transferable
Belief Model (TBM) \citep{smets1994transferable}.  In fact, the conjunctive rule is equivalent
to the Dempster rule  without the normalization
process. It has a fast and clear convergence towards a solution. But this rule has a strong assumption that  all the sources are reliable.
In real applications, it is difficult to be either satisfied or verified.  Moreover, the more sources there are,
the more chance that there is some unreliable evidence.}

{Smets \cite{smets1990combination} reasoned} that the mass on the empty set can play the role of alarm. When the global conflict (the mass assigned to the empty set) is
high, it indicates that there is  strong disagreement among the sources of mass functions to combine.  However, as
observed in \cite{martin2008conflict,liu2006analyzing,destercke2013toward}, the mass on the empty set is not sufficient to exactly describe the
conflict since it includes an amount of auto-conflict \cite{Martin06a}.
Sometimes when there is only a small amount of concordant evidence, the total conflict mass function, {\em i.e.} $m(\emptyset)=1$ will be an absorbing element.
Consequently, when combining a large number of (incompatible) mass functions using the conjunctive
rule, the global conflict  may tend to 1. This makes it impossible to reveal the
cause of high global conflict. We do not know whether it is due
to the  sources to fuse or  caused by the absorption power of the empty set \cite{martin2008conflict,lefevre2013preserve}. In other words, even
the combined mass function by the conjunctive rule is $m(\emptyset)\approx 1$, the proposition that the sources are highly conflicting may
be incorrect.

{In order to  rectify the drawbacks of the classical Dempster's rule and {Smets'} conjunctive rule,
many approaches have been made through the modification of the combination rule.
Some authors tried to find alternative repartitions of the conflict. A plethora of combination rules have been brought forward in this way. For example, \citet{yager1987dempster}
and \citet{dubois1988representation} suggested assigning the highly conflicting mass to the whole set or a particular set.
{The Proportional Conflict Redistribution (PCR) rule, which can distribute the partial conflicts
among the involved focal elements rather than to their
union, is developed in \cite{Martin06a,ilin2015information}.}
Apart from these approaches working directly on the combination rule, some studies manage the conflict
through evidence discounting, where the
reliability of sources is automatically and adaptively taken into
account \cite{dubois1988representation,martin2008conflict,martin2008general,zhao2016novel}.}

Most of the existing combination rules are not efficient when applied on a large number of sources
{due to} the ineffective way to handle
conflict or the high complexity of the computation. \citet{Orponen90a} proved that the
complexity of the conjunctive rule is NP-hard, but the complexity depends on the way to program the belief functions~\cite{daSilva92a}.
Some rules can manage efficiently the conflict but
have large complexity \cite{Martin06a,martin2007toward,smets97alpha,dubois1988representation}, making them infeasible when
applied to combine a large number of mass functions.

In this paper, a conjunctive-based combination rule, named {$\LNS$ (Large Number of Sources)}, is proposed  to
aggregate  a large number of mass functions. Our perspective on belief function
combination is that combining mass functions from different sources is
similar to combining opinions from multiple stake-holders in
group decision-making \cite{leung2013integrated}, {\em i.e.} the more one's opinion is consistent with the other experts, the more reliable the source is.
We assume that all the mass functions available are separable mass functions, which means they can
be expressed by a group of simple support mass functions. In many applications, the mass
assignments are directly in the form of Simple Support Functions (SSF) \cite{ds2}.
The advantage of SSFs is that we can group the mass functions in such a way that
sources in the same group share the same viewpoint. Mass functions in each small group
are first fused and then discounted according to the proportions. After
that the number of mass functions
participating the next global combination process is independent of the number of sources, but only depends on the number of
classes. As a result, the problem brought by the absorbing element (the empty set) using the conjunctive rule can be avoided. Moreover,
an approximation method when the number of mass functions is large enough is presented.
{The main contributions of this paper are as follows:}
\begin{itemize}
  \item A new conjunctive-based combination rule, named $\LNS$ rule, is brought froward. The property to reinforce the belief on the focal elements with which most of the sources agree is preserved in the proposed rule;
  \item {The assumption of the $\LNS$ rule on the reliability of the sources is more relaxed, as it does not require all the sources are reliable, but only at least half of them are reliable.}
  \item  {$\LNS$ can be used} to combine mass functions from a large number of sources, especially can be used to elicit the major opinion;
  \item {Derivation that the $\LNS$ rule is within acceptable complexity.}
\end{itemize}

The rest of this paper is organized as follows. In Section 2, some basic knowledge of belief function theory is briefly introduced. The proposed
evidence combination approach is presented in detail in Section 3.  Numerical examples are employed to compare different
combination rules and show the effectiveness of $\LNS$ rule in Section 4.
Finally, Section 5 concludes the paper.

\section{Background}
\subsection{Basic knowledge of belief function theory}
Let $\Theta=\{\theta_{1},\theta_{2},\ldots,\theta_{n}\}$ be  the discernment frame. A mass function is defined on the power
set $2^{\Theta}=\{A:A\subseteq\Theta\}$. The mass function $m:2^{\Theta}\rightarrow[0,1]$ is said to be a Basic Belief
Assignment (bba) on $\text{2}^{\Theta}$, if it satisfies:
\begin{equation}
\sum_{A\subseteq\Theta}m(A)=1.
\end{equation}
Every $A\in2^{\Theta}$ such that $m(A)>0$ is called a focal element, and the set of focal elements is denoted by $\mathcal{F}$. In a practical way of programming, the element of $2^\Theta$ can be arranged by natural order \cite{smets2002application}: $\theta_1, \theta_2, \{\theta_1, \theta_2\}, \theta_3, \cdots, \{\theta_1,\theta_2,\theta_3\}, \theta_4, \cdots, \Theta.$

The frame of discernment can also be a focal element.  If $\Theta$ is a focal element, the mass function is called non-dogmatic. The mass assigned to the frame of discernment, $m(\Theta)$, is interpreted as a degree of ignorance. In the case of total ignorance, $m(\Theta)=1$.
This type of mass assignment is  vacuous.
If there is only one focal element, {\em i.e.}
$m(A)=1, A \subset \Theta$, the mass function is  categorical.  Another special case of assignment is named consonant mass functions, where the focal
elements include each other as a subset, {\em i.e.} if $A, B \in \mathcal{F}, A \subset B ~\text{or}~ B \subset A$.

The credibility and plausibility functions are derived from a bba $m$ as in Eqs.~\eqref{bel} and \eqref{pl}:
\begin{equation}
Bel\text{(}A\text{)}=\sum_{B\subseteq A, B \neq \emptyset} m\text{(}B\text{)},~~\forall A\subseteq\Theta,
\label{bel}
\end{equation}
\begin{equation}
 Pl\text{(}A\text{)}=\sum_{B\cap A\neq\emptyset}m\text{(}B\text{)},~~\forall A\subseteq\Theta.
 \label{pl}
\end{equation}
Each quantity $Bel(A)$   measures the minimal belief on $A$ justified by available information on
$B(B \subseteq A)$ , while $Pl(A)$ is the maximal belief on $A$ justified by information on $B$
which are not contradictory with $A$ ($A \cap B \neq \emptyset$). 
The commonality function $q$ and the implicability function $b$ are defined  respectively as
\begin{equation}\label{qfun}
  q(A) = \sum_{A\subseteq B}m(B), ~~\forall A\subseteq\Theta
\end{equation}
and
\begin{equation}\label{bfun}
{ b(A) = Bel(A) + m(\emptyset),~~\forall A\subseteq\Theta.}
\end{equation}
A bba $m$ can be recovered from any of these functions. For instance,
\begin{equation}\label{qtom}
  m(A) = \sum_{B \supseteq A} (-1)^{|B|-|A|}q(B),~~\forall A\subseteq\Theta
\end{equation}
and \begin{equation}\label{btom}
  m(A) = \sum_{B \subseteq A} (-1)^{|A|-|B|}b(B),~~\forall A\subseteq\Theta.
\end{equation}

Belief functions can be transformed into a probability function by {Smets' method \cite{smets2005decision}}, where each mass
of belief $m(A)$ is equally distributed among the elements of $A$. This leads to the concept of pignistic probability, $\BetP$.
For all $\theta_i \in \Theta$, we have
\begin{equation}
 \label{pig}
	\BetP(\theta_i)=\sum_{A \subseteq \Theta | \theta_i \in A} \frac{m(A)}{|A|(1-m(\emptyset))},
\end{equation}
where $|A|$ is the cardinality of set $A$ (number of elements of $\Theta$ in $A$).
{Pignistic probabilities can help make a decision.}

\subsection{Consistency of mass assignments}

The consistency between two bbas can be defined in two different ways. Suppose the sets of focal elements for $m_1$ and $m_2$ are   $\mathcal{F}_1$
and $\mathcal{F}_2$ respectively.  Mass functions $m_1$ and $m_2$ are called strong consistent
if and only if
\begin{equation}
   \cap_{E\in \{\mathcal{F}_1 \cup \mathcal{F}_2\}} \neq \emptyset.
\end{equation}
Meanwhile, bbas $m_1$ and $m_2$ are called weak consistent if and only
if \begin{equation}
\forall A \in \mathcal{F}_1, B \in \mathcal{F}_2, A \cap B \neq \emptyset.
\end{equation}

Strong consistent evidence means that there is at least one element that is common to all
subsets \cite{sentz2002combination}. It is easy to see that, when $m_1$ and $m_2$ are strong consistent, they are
sure to be weak consistent. {This is the definition of consistency between belief functions.} The inconsistency within
an individual mass assignment can be defined similarly \cite{destercke2013toward}.

\subsection{Reliability-based discounting}
\label{reliability}

When the sources of evidence are not completely reliable, the discounting operation proposed by~\citet{ds2} and justified by~\citet{Smets93a}
{could} be applied. Denote the reliability degree of mass function $m$ by $\alpha \in[0,1]$, then the discounting operation can be defined as:
{\begin{equation}
  m^{'}(A) = \begin{cases}
    \alpha \times m(A) & \forall A \subset \Theta,\\
    1 - \alpha + \alpha \times m(\Theta) & \text{if}~ A = \Theta.
  \end{cases}
\end{equation}}
If $\alpha = 1$, the evidence is completely reliable and the bba will remain unchanged. On the contrary, if $\alpha=0$, the evidence is completely
unreliable. In this case the so-called vacuous belief function, $m(\Theta)=1$, {could} be got. It describes {the} total ignorance.

Before evoking the discounting process, the reliability of each sources should be known.
One possible way to estimate the reliability  is to use  confusion matrices~\cite{martin2005comparative}. 
Generally, the goal of discounting  is to reduce global conflict before combination. One can assume that
the conflict comes from the unreliability of the sources. Therefore, the source reliability estimation  is to some extent
linked to the estimation of conflict between sources.

%

Hence, \citet{martin2008conflict} proposed to use
a conflict measure to evaluate  the relative reliability of experts.
Once the degree of conflict is computed, the relative
reliability of the source can be computed accordingly. Suppose there are $S$ sources, $\mathcal{S}=\{s_1,s_2,\cdots,s_S\}$,
the  reliability discounting factor $\alpha_j$ of source $s_j$ can be defined as follows:
\begin{equation} \label{reliablitydiscount}
  \alpha_j = f\left(\conf\left(s_j, \mathcal{S}\right)\right),
\end{equation}
where 
$\conf\left(s_j, \mathcal{S}\right)$ quantifies the degree that source $s_j$ conflicts with the other sources in $\mathcal{S}$, and $f$ is a decreasing function.   The following function is suggested by the authors:
\begin{equation}
  \alpha_j = \left(1-\conf\left(s_j, \mathcal{S}\right)^\lambda\right)^{\frac{1}{\lambda}},
\end{equation}
where $\lambda > 0$.

In \cite{samet2013reliability}, the authors  considered to use those
two possible conflict origins, extrinsic measure and intrinsic measure, to estimate reliability. In their opinion, conflict may
not only come from the source's contradiction (extrinsic measure), but also from the confusion rate of a source (intrinsic measure).
The reliability discounting factor, called Generic Discounting Factor (GDF), is then suggested to be a weighted sum of the two items:
\begin{equation}
  \alpha = \frac{k \delta + l \beta}{k+l},
\end{equation}
where $k>0, l>0$ are the weight factors. In the above equation, $\delta$ denotes the internal conflict measure of the treated
source indicating its confusion rate while $\beta$ is the average
distance between the treated sources $s_i$ and $s_j$ where \linebreak $j\in \mathcal{S}, j \neq i$. Different intrinsic
and extrinsic conflict measures can be adopted here.

There are some other methods to estimate the reliability. {In \cite{schubert2011conflict1}, the authors proposed to
estimate the reliability of sources based on a degree of falsity. The bbas are sequentially and incrementally discounted
until the mass assigned to the empty set is smaller than a given threshold $k$. After that the discounted mass functions can be combined
using the conjunctive rule since there is little global conflict at this time.} In \cite{elouedi2001evaluation}, the source reliability is
obtained by  minimizing the distance between the pignistic probabilities computed from the discounted beliefs and the actual
value of the data. In \citet{samet2015reliability}, the authors proposed two different versions of generic discounting approaches:
weighted GDA and exponent GDA.
A new degree of disagreement  is proposed by \citet{yang2013discounted}, where the reliability discounting factor can be generated.
\citet{klein2011singular}  viewed the degree of
conflict as a function of discounting rates and introduced  a new  criterion assessing bbas' reliability. These reliability
estimation methods  either consider the distance (or dissimilarity) between each pair of bbas, or the mass assigned to the
empty set after the conjunctive combination. However, these methods are of high
complexity and not suitable for large data applications.

\subsection{Simple support function}
Suppose $m$ is a bba defined on the frame of discernment $\Theta$. If there exists a subset $A \subseteq \Theta$ such that $m$ could be expressed in the following form:
\begin{equation}
  m(X) = \begin{cases}
    w  & X = \Theta, \\
    1 - w & X = A, \\
    0 & \text{otherwise}.
  \end{cases}
\end{equation}
where $w \in \left[0,1\right]$, then the belief function related to bba $m$ is called a Simple Support Function (SSF) (also called simple mass function) \cite{ds2} focused on $A$. Such a SSF can be denoted by $A^w(\cdot)$ where the exponent $w$ of the focal element $A$ is the basic belief mass (bbm)  given to the frame of discernment $\Theta$, $m(\Theta)$. The complement of $w$ to 1, {\em i.e.} $1-w$, is the bbm allocated to $A$ \cite{smets1995canonical}. If $w=1$ the mass function represents the total ignorance, if $w=0$ the mass function is a categorical bba on $A$.

A belief function is separable if it is a SSF or if it is the conjunctive combination of some SSFs \cite{denoeux2008conjunctive}. In the work of \cite{denoeux2008conjunctive}, this kind of separable masses is called u-separable where ``u" stands for ``unnormalized", indicating the conjunctive rule is the unnormalized version of Dempster-Shafer rule.
The set of separable mass functions is not obvious
to obtain.
It is easy to see consonant mass functions (the focal element are nested) are separable~\cite{Ke14a}. \citet{smets1995canonical} defined the Generalized Simple Support Function (GSSF) by relaxing the weight $w$ to $[0,\infty)$.
Those GSSFs with $w\in (1,\infty)$ are called Inverse Simple Support Functions (ISSF). {Smets proved} all non-dogmatic mass functions are separable if one uses GSSFs. For any non-dogmatic belief function $m_0$, the canonical  decomposition method proposed by Smets is as
follows. First, calculate the
commonality number for all focal elements, which is given by
\begin{equation}
  Q_0(X) = \sum_{B \supseteq X} m_0(B).
\end{equation}
Secondly for any $A \subseteq \Theta$, calculate $w_A$ value as follows:
\begin{align}
  w_A= \prod_{X\supseteq A} Q_0(X)^{(-1)^{|X|-|A|+1}}.
\end{align}
Then the belief function $m_0$ can be represented by the conjunctive combination of all the functions $A_{w_A}$, {\em i.e.}
\begin{equation}
  m_0 = \ocap_{A \subseteq \Theta} A^{w_A},
\end{equation}
where $\ocap$ denotes the conjunctive combination rule.  For fast computation, the  {Fast M{\"o}bius Transform (FMT) method}
 \cite{kennes1992computational} can be evoked.

\subsection{Some combination rules}
How to combine efficiently several bbas coming from distinct sources  is a major information fusion problem in the belief function
framework. Many rules have been proposed for such a task. 
Here we just briefly recall how some most popular rules are mathematically defined.

When information sources are reliable, the used fusion operators can be based on the conjunctive combination.
If bbas $m_j, j=1,2,\cdots,S$ describing $S$
distinct items of evidence on $\Theta$, the included result of the \textbf{conjunctive rule} \cite{smets1994transferable} is defined as
\begin{equation}
\label{conjunctive}
m_\conj(X) =(\ocap_{j=1, \cdots, S}m_j)(X)= \sum\limits_{Y_1 \cap \cdots \cap Y_S = X} \prod_{j=1}^{S}m_j(Y_j),
\end{equation}
where 
$m_j(Y_j)$ is the mass allocated to $Y_j$ by expert $j$.
To apply this rule, the sources are assumed reliable and cognitively independent.

Another kind of conjunctive combination is  \textbf{Dempster's rule} \cite{dempster1967upper}. Assuming that 
$m_\conj(\emptyset) \neq 1$, the result of the combination by Dempster's rule is
\begin{equation}\label{ds}
   m_\DS(X)= \begin{cases}
    0 & \text{if}~ X = \emptyset,\\
     \frac{m_\conj(X)}{1-m_\conj(\emptyset)} & \text{otherwise}.
  \end{cases}
\end{equation}
The item
$$\kappa \triangleq m_\conj (\emptyset)=\sum\limits_{Y_1 \cap \cdots \cap Y_S = \emptyset} \prod_{j=1}^{S}m_j(Y_j)$$ is generally called Dempster's
degree of conflict  of the combination or the inconsistency of the combination. As the conjunctive
rule is not idempotent, $m_\conj (\emptyset)$ includes an amount of auto-conflict \cite{osswald2006understanding}, and it is called global conflict to
make the difference.

The conjunctive rule can be applied  only if all the experts are
reliable. In the other case, the \textbf{disjunctive rule} \cite{smets1993belief}, which
only assumes that at least one of the sources is reliable, can be used. The
disjunctive combination of $S$ sources can be defined as
\begin{equation}
\label{disjunctive}
m_\disj(X) =\left(\ocup_{j=1, \cdots, S}m_j\right)(X)= \sum\limits_{Y_1 \cup \cdots \cup Y_S = X} \prod_{j=1}^{S}m_j(Y_j).
\end{equation}

The conjunctive and disjunctive rules can be conveniently expressed by
means of the commonality function $q$ (Eq.~\eqref{qfun}) and the
implacability function $b$ (Eq.~\eqref{bfun}) \cite{smets1993belief}. Let $q_i$
and $b_i$ be the commonality function and implacability function respectively (associated with $m_i$), then the commonality function
of the conjunctive combination of $S$ bbas is
\begin{equation}
  q_\conj(A) = \prod_{i=1}^S q_i(A), ~~\forall A\subseteq\Theta
\end{equation}
while the implacability function of the disjunctive combination of $S$ bbas is
\begin{equation}
  b_\disj(A) = \prod_{i=1}^S b_i(A), ~~\forall A\subseteq\Theta.
\end{equation}

Since functions $m$, $q$ and $b$ (as
well as $bel$ and $pl$) are equivalent representations, the mass function $m$ can be recovered using
the Fast M{\"o}bius Transform (FMT) method given the functions $q$ and $b$. The conversion can be
done in time proportional to $n 2^n$ \cite{Wilson00a}\footnote{This
is based on the assumption that the mass functions
are arranged in natural order. If not, the complexity is proportional to $n^2 2^n$. The complexity analysis in this work all assumes that
the bbas to be combined are encoded using the  natural order.}. For the conjunctive combination of $S$ sources, the $S$ bbas should
be converted into commonality functions first. After calculating the product
of $S$ commonality functions, another transformation from $m$ to $q$ should be evoked. Overall the total complexity is
$O(Sn2^n+S2^n+n2^n)$, and the time needed is proportional to $S n 2^n$ \cite{Wilson00a,Denoeux02a}.

The conflict could be redistributed on partial ignorance like in the Dubois and Prade rule (\textbf{\bm{$\DP$} rule}) \cite{dubois1988representation},
which can be seen as a mixed conjunctive and disjunctive rule. For all $X \subseteq \Theta, X \neq \emptyset$:
\begin{align}
 m_\DP(X) = \sum_{Y_1 \cap \cdots \cap Y_S = X} \prod_{j=1}^S m_j(Y_j) + & \nonumber \\ \sum_{\mbox{\tiny $\begin{array}{c}Y_1\cup\cdots \cup Y_S = X \\ Y_1 \cap \cdots \cap Y_S = \emptyset\end{array}$}} \prod_{j=1}^S m_j(Y_j) ,
\end{align}
where $m_j$ is the mass function delivered by expert $j$. 
In a general case, this rule cannot be programmed  with the Fast M{\"o}bius Transform method because all the partial conflict must be considered. If the implementation is made like that in Ref.~\cite{Martin09a}, it takes much more time than the conjunctive rule.

{\citet{denoeux2008conjunctive} proposed} a family of conjunctive and disjunctive rules using triangular norms. The \textbf{cautious rule} \cite{denoeux2006cautious, chin2015weighted} belongs to that family and could be used to combine mass functions for which independence assumption is not verified. Cautious combination
of $S$ non-dogmatic mass functions $m_j, j=1,2,\cdots,S$  is defined by the bba with the following weight function:
\begin{equation}
  w(A)= \mathop{\wedge}\limits_{j=1}^S w_j(A), ~~ A \in 2^\Theta \setminus \Theta.
\end{equation}
We thus have
\begin{equation}
 m_\caut(X) = \ocap_{A \subsetneq \Theta} A^{\mathop{\wedge}\limits_{j=1}^S w_j(A)},
\end{equation}
where $A^{w_j(A)}$ is the simple support function focused on $A$ with
weight function $w_j(A)$ issued from the canonical decomposition of $m_j$.
Note also that $\wedge$ is the min operator. The time consumption of the cautious rule includes the canonical
decomposition of non-dogmatic mass functions and
is therefore bigger than the conjunctive rule.  If this rule is implemented in Fast M{\"o}bius Transform method, the complexity is proportional to
$Sn2^n$.

{\citet{murphy2000combining} presented} the \textbf{average combination rule}  and proposed to utilize
the mean of the basic belief assignments as the fusion of evidence. Therefore, for each focal element $X \in 2^\Theta$ of $S$ mass functions, the combined one is defined as follows:
\begin{equation}
 m_\Ave (X) = \frac{1}{S} \sum_{j=1}^S m_j(X), \forall X \subseteq \Theta.
\end{equation}
The complexity of the average is proportional to $S2^n$.

A  family of fusion rules based on new Proportional Conflict Redistributions (PCR) for the combination of uncertainty and conflicting information
have been developed in Dezert--Smarandache Theory (DSmT) framework  \cite{smarandache2006advances}. Among them, the fusion
rule called  \textbf{$\PCR$} proposed by \citet{Martin06a}  is one of the most popular one among the PCR rules. For the combination of
$S>2$ sources, the fused mass is given by $m_\PCR(\emptyset)= 0$, and for $X \neq \emptyset$ in $2^\Theta$
\begin{align}\label{pcr6equ}
  &m_\PCR(X) = m_\conj(X) + \sum_{i=1}^S \scalebox{1.2}{$\Bigg\{$}\left(m_i\left(X\right)\right)^2 \times \nonumber \\ &\sum_{\mbox{\footnotesize $\begin{array}{c}\mathop{\bigcap}\nolimits_{k=1}^{S-1} Y_{\sigma_i(k)}\cap X \equiv \emptyset \\ \left(Y_{\sigma_i(1)},\cdots, Y_{\sigma_i(S-1)}\right) \in \left(2^\Theta\right)^{S-1} \end{array}$}}\scalebox{0.9}{$\left( \frac{\prod\limits_{j=1}^{S-1} m_{\sigma_i(j)}\left(Y_{\sigma_i(j)}\right)}{m_i(X)+\sum\limits_{j=1}^{S-1}m_{\sigma_i(j)}\left(Y_{\sigma_i(j)}\right)}\right)$}
  \scalebox{1.25}{$\Bigg\}$},
\end{align}
where $\sigma_i$ counts from 1 to $S$ avoiding $i$:
\begin{equation}
  \begin{cases}
    \sigma_i(j) = j & \text{if}~~ j<i, \\
    \sigma_i(j) = j+1 & \text{if}~~ j\geq i .
  \end{cases}
\end{equation}
As $Y_i$ is a focal element of expert/source $i$, we have $m(Y_i)>0$. Then $$m_i(X)+\sum\limits_{j=1}^{S-1}m_{\sigma_i(j)}\left(Y_{\sigma_i(j)}\right) \neq 0.$$ In Eq.~\eqref{pcr6equ}, $m_\conj$ is the conjunctive rule given by Eq.~\eqref{conjunctive}. Here again, the Fast M{\"o}bius Transform method to program the belief functions is not generally the best way. If the implementation is made like that  in Ref.~\cite{Martin09a}, the time consumption is very high. 


\section{A combination rule for a large number of mass functions}

The main idea of the conjunctive combination rule is to reinforce the belief on the focal elements  with which most of the sources agree.
{\citet{martin2008conflict} showed that} the mass on the empty set, which is an absorbing
element, tends quickly to 1 with the number of sources when combining inconsistent bbas.  Consequently, when using Dempster
rule ({Eq.~\eqref{ds}}), the gap between $\kappa$ and
1 may rapidly exceed machine precision, even if the combination is valid theoretically. In that case the fused bba by the conjunctive rules (normalized
or not) and the pignistic probability are inefficient. Moreover,  the assumption that  all the sources are reliable for
the conjunctive combination rule is difficult to reach in real applications. The more sources there are, the less chance that this assumption is valid.

The principle of the conjunctive rule with the reinforcement of  belief and the role of the empty set as an alarm are  essential in the theory
of belief functions. In order to propose a rule which can be adapted to the combination of a large number of mass functions and keep
the previous behavior, the following assumptions are made:
\begin{enumerate}
\item[$\bullet$] The majority of sources are reliable;
 \item[$\bullet$] The larger extent one source is consistent with others, the more reliable the source is;
 \item[$\bullet$] The sources are cognitively  independent \cite{smets1993belief}.
\end{enumerate}
{These assumptions seem reasonable if we consider combing mass functions as some kind of group decision making problems.
As a result, the proposed  rule will give more importance to the groups of mass functions 
that are in a domain}, and it is without
auto-conflict~\cite{Martin06a,lefevre2013preserve}.
In order to take into account this effect, this rule will discount the mass functions according to the
number of sources giving bbas with the same focal elements. The discounting factor is directly given by the proportion of
mass functions with the same focal elements. This procedure is for the elicitation of the majority opinion.

The simple support mass functions are considered here. In this case,  the mass functions can be grouped in the light of their focal
elements (except the frame $\Theta$). To make the rule applicable on separable mass functions, 
the decomposition process should be performed to decompose  each bba into simple
support mass functions.  In most of applications, the basic belief can be defined using separable mass functions, such as simple support
functions~\cite{denoeux1995k} 
and consonant mass functions~\cite{Dubois90a,Aregui08a}.

Hereafter we describe the proposed $\LNS$ rule for simple support functions, and then an approximation calculation method
of $\LNS$ rule is suggested.

{\subsection{$\LNS$ rule for simple support functions}}
Suppose that each evidence  is represented by a SSF. Then all the bbas  can be divided into at most $2^n$ groups (where $n=|\Theta|$). It is easy to
see that there is no conflict at all in each group because of consistency. The focal elements of the SSF are singletons  and
$\Theta$ itself. For the combination of bbas inside each group, the conjunctive rule can be employed directly. Then the fused bbas are  discounted
according to the number of mass functions in each group. Finally, the global combination of the bbas of different groups is preformed also using the
conjunctive rule. Suppose  that all bbas are defined on the frame of
discernment $\Theta=\{\theta_1,\theta_2,\cdots,\theta_n\}$, and denoted by  $m_j = (A_i)^{w_j}, j=1,\cdots, S$ and $i=1,2,\cdots,c$, where $c\leq 2^n$. 
The detailed process of the combination is listed as follows. Our proposed rule called $\LNS$ for Large Number of Sources rule is composed of the four
following steps:
\begin{enumerate}
  \item Cluster the simple bbas into $c$ groups based on their focal element $A_i$. For the convenience, each class is labeled by
  its corresponding focal element.
  \item Combine the bbas in the same group. Denote the combined bba in group $A_k$ by SSF  $$\hat{m}_k= (A_k)^{\hat{w}_k}, k=1,2,\cdots,c.$$
 {Let the number of bbas in group $A_k$ is $s_k$. If the conjunctive rule is adopted,  we have
  \begin{equation}
   \hat{m}_k=\ocap_{j=1, \cdots, s_k}m_j=(A_k)^{\displaystyle \prod_{j=1}^{s _k} w_j}.
  \end{equation}}

  \item Reliability-based discounting. Suppose the fused bba of all the mass functions
  in $A_k$ is $\hat{m}_k$. At this time, each group can be regarded as a source, and there are $c$ sources in total. The reliability of one source can be estimated as compared to a group of sources. In our opinion, the reliability of source $A_k$ is related to the proportion
  of bbas in  this group. The larger the number of bbas in group $A_k$ is, the more reliable $A_k$ is.  Then the reliability discounting
  factor of $\hat{m}_k$ can be defined as:
      \begin{equation}
      \label{discountfactorSimple}
      \alpha_k = \frac{s_k}{\displaystyle  \sum_{i=1}^{c} s_i}.
      \end{equation}
 {In order to keep the mass function  representing total ignorance as a neutral element of the rule, in Eq.~\eqref{discountfactorSimple}
 we let $a_k=0$ for the group with $A_k=\Theta$.}
      Another version of the discounting can be given by a factor taking into account the precision of the group by:
      \begin{equation}
      \label{discountfactor}
      \alpha_k = \frac{\beta_k^\eta s_k}{\displaystyle \sum_{i=1}^c \beta_i^\eta s_i},
      \end{equation}
      where
      \begin{equation}
        \beta_k = \frac{|\Theta|}{|A_k|}.
      \end{equation}

      Parameter $\eta$ can be used to adjust the precision of the combination results.  The larger the value of $\eta$ is, the
      less imprecise the resulting bba is.
      The discounted bba of $\hat{m}_k$ can be denoted by SSF $\hat{m}_k^{'}=(A_k)^{\hat{w}_k^{'}}$ with $\hat{w}_k^{'} = 1 - \alpha_k + \alpha_k \hat{w}_k$. As we can see, when the number of bbas in one group is larger, $\alpha$ is closer to 1. That is to say, the fused mass in this group is more reliable.

  \item Global combine the fused bbas in  different groups using the conjunctive rule:
  \begin{equation}
  \label{lastcombination}
   m_\LNS=\ocap_{k=1, \cdots, c} \hat{m}_k'=\ocap_{k=1, \cdots, c} (A_k)^{\hat{w}_k^{'}}.
  \end{equation}

\end{enumerate}
\textbf{Remarks:}
\begin{enumerate}
\item[$\bullet$]The reliability estimation method proposed here is very simple compared with the previous mentioned methods in Section~\ref{reliability}, where usually the distance
      between bbas should be calculated or a special learning process is required. In the $\LNS$ rule, to evaluate the
      reliability discounting factor, we only need to count the number of SSFs in each
      group. Note that other  reliability estimation methods can  also be used here.
  \item[$\bullet$]In the last step of combination, as the number of mass functions that take part in the global combination is small (at most $2^n$), other combination rules such as $\DP$ rule and PCR rules are also possible in practice instead of Eq.~\eqref{lastcombination}.
\end{enumerate}

{\subsection{$\LNSA$ rule for the approximated combination}}
If there is a large number of mass functions in each group, an approximation method is suggested here
to calculate the combined mass in the given group.
Suppose the mass functions in group with focal element $A_k$ ($k=1,2,\cdots, c$) are: 
\begin{equation}
  m_{j}(A)=\begin{cases}
    1-w_{j} & A = A_k,\\
    w_{j} & A = \Theta, \\
    0  & \text{otherwise},\\
  \end{cases} ~ 0\leq w_{j}<1, j =1,2,\cdots, s_k.
\end{equation}
The combination of the masses in this group using the conjunctive rule is
\begin{equation}
  \hat{m}_k(A)= \begin{cases}
    1-\prod\limits_{j=1}^{s_k}w_{j} & A = A_k,\\
    \prod\limits_{j=1}^{s_k}w_{j} & A = \Theta,\\
   0  & \text{otherwise}.\\
  \end{cases}
\end{equation}
It is easy to get
\begin{equation}
  \lim_{s_k\rightarrow \infty}\hat{m}_k(A)= \begin{cases}
    1 & A = A_k,\\
    0 & A = \Theta,\\
   0  & \text{otherwise}.
  \end{cases}
\end{equation}
This is an illustration of the conjunctive property. After the discounting with factor $\alpha_k$, the fused bba using for the global combination is
\begin{equation}
    \lim_{n_k\rightarrow \infty}\hat{m}^{'}_k(A)= \begin{cases}
    \alpha_k & A = A_k,\\
    1-\alpha_k & A = \Theta,\\
   0  & \text{otherwise}.
  \end{cases}
\end{equation}
It can be represented by SSF
\begin{equation}
   \hat{m}_k^{'} = (A_k)^{1-\alpha_k},
\end{equation}
where $\alpha_k$ is shown in Eq.~\eqref{discountfactorSimple} or~\eqref{discountfactor}.
If the conjunctive rule is adopted for the global combination at step 4, the final bba we get is
\begin{equation}
  m_{\LNSA} = \ocap (A_k)^{1-\alpha_k}.
\end{equation}

In this approximate rule for the large number of sources, {the initial mass functions
is no longer considered, and the combination process of the bbas inside each group  is not required any more. This
can accelerate the algorithm to a large extent. The $\LNS$ and $\LNSA$ rule provide different results when the number of sources is small.
However, when the number of sources is large enough, they can be regarded as equivalent. }

\subsection{Properties}
\label{secproperties}
The proposed rule is commutative, but not associative. The rule is not idempotent, but there is no absorbing element. 
The vacuous mass function is a neutral element of the $\LNS$ rule. 

There are four steps when applying $\LNS$ rule\footnote{The source code for $\LNS$ rule can be found in R package \textit{ibelief} \cite{ibelief}.}:
decomposition (not necessary for simple support mass functions), inner-group combination, discounting and global combination. The $\LNS$ rule has the
same memory complexity as some other rules such as conjunctive, $\DS$ and cautious rules if all the rules are combined globally
using FMT method. Only $\DP$ and $\PCR$ rules have higher memory complexity because of the partial conflict to manage. Suppose the number of
mass functions to combine is $S$, and the number of elements in the frame of discernment  is $n$. The complexity for decomposing\footnote{In the
decomposing process, the Fast M{\"o}bius Transform method is used.
 }
mass functions to SSFs is $O(Sn2^n)$.
For combining the mass functions in each group, due to the structure of the simple support mass functions, we only need to
calculate the product of the masses on only one
focal element $\Theta$. Thus the complexity is $O(S)$. The complexity of the discounting is $O(2^n)$. In the
process of global combination, the bbas are all SSFs. If we use  the Fast M{\"o}bius Transform method
, the complexity is $O(n2^n)$. And there are at most $2^n$ mass functions participating  the following discounting and global
conjunctive combination processes. Since in most application cases with a large number of mass functions, we have $2^n \ll S$, the last  two steps are
not very time-consuming. The total complexity of $\LNS$ is $O(Sn2^n+S+2^n+n2^n)$ and so is {approximately} equivalent to $O(Sn2^n)$.

For the approximate method, we can also save the time for inner combination and the discounting. The fused mass in each group is calculated by the proportions, and the complexity is also $O(S)$. Although the approximate method does not reduce the complexity, in the experimental part, we will show that it will save some running time in applications when $S$ is quite large.

We remark here that one of the assumptions of $\LNS$ rule is that the majority of sources are reliable.  However, this condition  is
not always satisfied in every applicative context. Consider here an example with two sensor
technologies: TA and TB. The system has two TA-sensors ($S_1$ and $S_2$),  and one TB-sensor $S_3$. Suppose also a
parasite signal causes TA sensors to malfunction. In this situation, the majority of sensors are unreliable. And we could not
get a good result if the $\LNS$ rule is used directly as $\LNS(S_1,S_2,S_3)$ at this time.
Actually there is an underlying hierarchy in the sources of information,  $\LNS$
rule could be evoked according to  the hierarchy, such as $\LNS(\LNS(S_1,S_2),S_3)$. We will study that more in the future work.

\section{Experiments}
\label{Experiments}
In this section, several experiments will be conducted to illustrate the behavior of the proposed combination rule $\LNS$ and
to compare  with other classical rules. 
Some different types of randomly generated mass functions  will be used. The function \textit{RandomMass} in R package \textit{ibelief}
\cite{ibelief} is adopted to generate random mass functions \cite{burger2013randomly}.


\Exp \change{~(Elicitation of the majority opinion).} {In some applications, the elicitation of the majority opinion is very important. 
In this experiment, {it is assumed that reliable sources can provide some imprecise} and uncertain information, which is  assumed to be  in the form of the mass
functions $m_j$ $(j = 1, 2, \cdots, 6)$ over the same discernment frame $\Theta=\{\theta_1,\theta_2,\theta_3\}$:}
\begin{align*}
& m_1: m_1(\{\theta_1\}) = 0.12, ~m_1(\Theta)=0.88,\nonumber \\
& m_2: m_2(\{\theta_1\}) = 0.16, ~m_2(\Theta)=0.84,\nonumber \\
& m_3: m_3(\{\theta_1\}) = 0.15, ~m_3(\Theta)=0.85,\nonumber \\
& m_4: m_4(\{\theta_1\}) = 0.11, ~m_4(\Theta)=0.89,\nonumber\\
& m_5: m_5(\{\theta_1\}) = 0.14, ~m_5(\Theta)=0.86,\nonumber\\
& m_6: m_6(\{\theta_2\}) = 0.95, ~m_6(\Theta)=0.05. \nonumber\\
\end{align*}

\vspace{-1.6em}
As can be seen, the first five sources share similar belief (supporting $\{\theta_1\}$)
whereas the sixth one delivers a mass function strongly committed to another solution (supporting $\{\theta_2\}$).
These six mass functions cannot be regarded as conflicting, because the majority of evidence shows the preference of $\{\theta_1\}$.
{Here, source 6, is assumed not reliable} since it contradicts with all the other sources.

The combination results by conjunctive rule, $\DS$ rule, disjunctive rule, $\DP$ rule, $\PCR$ rule, cautious rule, average rule
and the proposed $\LNS$ rule\footnote{As the focal elements are singletons except $\Theta$, parameter $\eta$ has no effects on
the final results when using $\LNS$ rule.} are depicted in Table
\ref{6massEx1}. As can be observed, the conjunctive rule assigns most of the belief to the empty set, regarding the sources as highly
conflictual. $\DS$ rule, $\DP$ rule, $\PCR$ rule and average rule redistribute
all the global conflict to other focal elements. {The disjunctive rule gives the total} ignorance mass functions.
{The cautious rule and the proposed}  $\LNS$ rule keep some of the conflict and redistribute the remaining.
But the belief given to $\{\theta_2\}$ is more than that to $\{\theta_1\}$
when using  $\DS$, $\DP$, $\PCR$, cautious
and the average rules, which indicates that these rules are not robust to the unreliable evidence. The obtained fused bba by the proposed rule
assigns  the largest  mass to focal element $\{\theta_1\}$, which is
consistent with the intuition. It keeps a certain level of global conflict, and  at
the same time reflects the superiority of $\{\theta_1\}$ compared with $\{\theta_2\}$. From the results we can see that only
the $\LNS$ rule can correctly elicit the major opinion.

\begin{table*}[ht]
\centering \caption{The combination of six masses. For the names of columns, $\theta_{ij}$ is used to denote $\{\theta_i, \theta_j\}$.}
\resizebox{\textwidth}{!}{
\begin{tabular}{lllrrrrrrr}
  \hline
 & Conjunctive & $\DS$ & Disjunctive & DP & PCR6 & Cautious & Average & $\LNS$ \\
  \hline
$\emptyset$ & 0.49313 & 0.00000 & 0.00000 & 0.00000 & 0.00000 & 0.15200 & 0.00000 & 0.06849 \\
  $\{\theta_1\}$ & 0.02595 & 0.05120 & 0.00000 & 0.02595 & 0.04783 & 0.00800 & 0.11333 & 0.36408 \\
  $\{\theta_2\}$ & 0.45687 & 0.90136 & 0.00000 & 0.45687 & 0.56639 & 0.79800 & 0.15833 & 0.08984 \\
  $\{\theta_{1},\theta_{2}\}$ & 0.00000 & 0.00000 & 0.00004 & 0.49313 & 0.00000 & 0.00000 & 0.00000 & 0.00000 \\
  $\{\theta_3\}$ & 0.00000 & 0.00000 & 0.00000 & 0.00000 & 0.00000 & 0.00000 & 0.00000 & 0.00000 \\
  $\{\theta_{1},\theta_{3}\}$ & 0.00000 & 0.00000 & 0.00000 & 0.00000 & 0.00000 & 0.00000 & 0.00000 & 0.00000 \\
  $\{\theta_{2},\theta_{3}\}$ & 0.00000 & 0.00000 & 0.00000 & 0.00000 & 0.00000 & 0.00000 & 0.00000 & 0.00000 \\
  $\Theta$ & 0.02405 & 0.04744 & 0.99996 & 0.02405 & 0.38578 & 0.04200 & 0.72833 & 0.47759 \\
   \hline
\end{tabular}
}
\label{6massEx1}
\end{table*}

{The $\LNS$ rule is a} conjunctive based combination rule for mass functions with different reliability degrees. As mentioned before,
the principle of the $\LNS$ rule is similar that of Schubert's method  \cite{schubert2011conflict1}. Table \ref{6massesSh} lists the results
by Schubert's combination method with  different values of $k$. As can be seen,
the result by the use of the $\LNS$ rule is similar to that by Schubert's method with a small value of
threshold $k$. When $k$ is set small, the discounting process in Schubert's
method needs more steps. And in each step, the conjunctive rule should be evoked to calculate the falsity. It is more complex compared
with the reliability estimation process of the $\LNS$ rule in that sense.

\begin{table}[ht]
\centering \caption{The combination of six masses by Schubert's {method} with different values of $k$.}
\begin{tabular}{lrrrrrr}
  \hline
 $k$ & 0.1 & 0.2 & 0.3 & 0.4 & 0.5 \\
  \hline
  $\emptyset$ & 0.09776 & 0.19471 & 0.28680 & 0.37803 & 0.46444 \\
  $\{\theta_1\}$ & 0.32187 & 0.26219 & 0.19350 & 0.12081 & 0.04980 \\
  $\{\theta_2\}$ & 0.13521 & 0.23145 & 0.31033 & 0.37979 & 0.43871 \\
  $\{\theta_{1},\theta{2}\}$ & 0.00000 & 0.00000 & 0.00000 & 0.00000 & 0.00000 \\
  $\{\theta_3\}$ & 0.00000 & 0.00000 & 0.00000 & 0.00000 & 0.00000 \\
  $\{\theta_{1},\theta_{3}\}$ & 0.00000 & 0.00000 & 0.00000 & 0.00000 & 0.00000 \\
  $\{\theta_{2},\theta{3}\}$ & 0.00000 & 0.00000 & 0.00000 & 0.00000 & 0.00000 \\
  $\Theta$ & 0.44516 & 0.31165 & 0.20937 & 0.12137 & 0.04704 \\
   \hline
\end{tabular}\label{6massesSh}
\end{table}

We also compare with another reliability discounting based combination method proposed by \citet{martin2008conflict}. Same
as Schubert's method, after the reliability degree
of each source is estimated, the bbas are discounted following with a conjunctive combination.
There is a parameter $\lambda$ in the method to adjust the discounting factor. The results varying with
different values of $\lambda$ are shown in Table \ref{6massesMar}.  We can see this rule is similar to $\LNS$ rule when $\lambda$ is set to be
around 1. When $\lambda$ is not well set, the results are not good. Moreover, \change{in this method, the distance between
bbas should be calculated first. Consequently, it increases the complexity and makes  the method not  feasible for
combining a large number of sources.}

\begin{table}[ht]
\centering \caption{The combination of six masses by Martin's {method} with different values of $\lambda$.}
\begin{tabular}{lrrrrrrr}
  \hline
 $\lambda$ & 0.1 & 0.5 & 1 & 1.5 & 2 \\
  \hline
$\emptyset$ & 0.00000 & 0.00350 & 0.10485 & 0.23330 & 0.31956 \\
  $\{\theta_1\}$ & 0.00000 & 0.21206 & 0.34700 & 0.26789 & 0.19410 \\
  $\{\theta_2\}$ & 0.00000 & 0.01272 & 0.12719 & 0.23219 & 0.30256 \\
  $\{\theta_{1},\theta_{2}\}$ & 0.00000 & 0.00000 & 0.00000 & 0.00000 & 0.00000 \\
  $\{\theta_3\}$ & 0.00000 & 0.00000 & 0.00000 & 0.00000 & 0.00000 \\
  $\{\theta_{1},\theta_{3}\}$ & 0.00000 & 0.00000 & 0.00000 & 0.00000 & 0.00000 \\
  $\{\theta_{2},\theta_{3}\}$ & 0.00000 & 0.00000 & 0.00000 & 0.00000 & 0.00000 \\
  $\Theta$ & 1.00000 & 0.77172 & 0.42096 & 0.26661 & 0.18378 \\
   \hline
\end{tabular}\label{6massesMar}
\end{table}

\Exp \change{~(The discounting mechanism).} {In this experiment, we will discuss the reliability discounting mechanism of the $\LNS$ rule.
Two reliability discounting methods proposed by  \citet{schubert2011conflict1} and \citet{martin2008conflict} will be used to compare. Same as the
$\LNS$ rule, after the discounting process by these two methods, the conjunctive rule \change{is} adopted to combine the new mass functions. For simplicity, here we call the combination rule,
\change{where the Schubert's discounting method (or Martin's discounting method) is first evoked and then
the conjunctive combination rule is used}, ``Schubert's method" (Martin's method, correspondingly).}
A set of $3*x$ bbas on a frame  of discernment $\Theta = \{\theta_1, \theta_2\}$ are generated,  $x$ of them are
unreliable while $2*x$ are reliable. The reliable sources
assign a large mass to the singleton $\{\theta_1\}$. The unreliable sources assign a large mass to the singleton $\{\theta_2\}$. The gain
factor for sequential discounting in Schubert's method is set to be 0.1 here.
{Schubert and Martin's methods} are evoked with different values of $k$ and $\lambda$ respectively. Let $x=10$, the fused bbas
by the use of different rules are listed in Table \ref{exp3add}.

\begin{table*}[ht]
\centering \caption{The combination results by different rules.}
\resizebox{\textwidth}{!}{
\begin{tabular}{l|rrrr|rrrr|rrrr}
  \hline
  & \multicolumn{4}{c|}{Schubert's method} &\multicolumn{4}{c|}{{Martin's method}}&\multicolumn{1}{c}{$\LNS$}\\
  \cline{2-9}
 & $k=0.2$ & $k=0.3$ & $k=0.5$ & $k=0.7$ & $\lambda=0.3$ & $\lambda=0.4$ & $\lambda=0.6$ & $\lambda=1$ &  \\
  \hline
  $\emptyset$ & 0.19949 & 0.29860 & 0.49704 & 0.69306 & 0.00248 & 0.10019 & 0.60681 & 0.98649 & 0.15060 \\
  $\{\theta_1\}$ & 0.80051 & 0.70140 & 0.50296 & 0.30694 & 0.16901 & 0.56713 & 0.38729 & 0.01351 & 0.48612 \\
  $\{\theta_2\}$ & 0.00000 & 0.00000 & 0.00000 & 0.00000 & 0.01200 & 0.04995 & 0.00360 & 0.00000 & 0.08593 \\
  $\Theta$ & 0.00000 & 0.00000 & 0.00000 & 0.00000 & 0.81650 & 0.28274 & 0.00230 & 0.00000 & 0.27735 \\
   \hline
\end{tabular}}\label{exp3add}
\end{table*}

From the table we can see, the behavior of Martin's discounting method is similar to
that of $\LNS$ rule when $\lambda$ is set around 0.4. {The conjunctive combination \change{based on} Schubert's discounting
does not give any belief to $\{\theta_2\}$  and $\Theta=\{\theta_1, \theta_2\}$ at all although there are  $1/3$ of sources supporting
$\{\theta_2\}$. Moreover, when $k$ is larger, most of the mass is assigned to the empty set in this rule. From \change{these} results we can see that
    only $\LNS$ rule can give more belief on $\{\theta_1\}$ which can be regarded as the major opinion.} The time elapsed for Schubert's {method} with different values of threshold $k$ is listed in Table \ref{Exp3addsh}. The smaller
the value of $k$ is, the more discounting  steps are required in Schubert's method. Consequently, the time consumption becomes larger.
\change{The running time for both $\LNS$ rule and Martin's method is less than one second.} Schubert's method is much more time-consuming.
\begin{table*}[ht]
\centering \caption{Time elapsed for Schubert's {method} with different values of $k$.}
\begin{tabular}{rrrrrrrrrr}
  \hline
 & 1 & 2 & 3 & 4 & 5 & 6 & 7 & 8 & 9 \\
  \hline
$k$ & 0.10 & 0.20 & 0.30 & 0.40 & 0.50 & 0.60 & 0.70 & 0.80 & 0.90 \\
 Time Elapsed (s)  & 46.81 & 21.64 & 13.46 & 9.28 & 6.64 & 4.88 & 3.67 & 2.73 & 1.79 \\
   \hline
\end{tabular}\label{Exp3addsh}
\end{table*}

We have also tested the combination methods based on the discounting factors proposed by \citet{schubert2011conflict1} and \citet{martin2008conflict}
on \change{some simple support mass functions with arbitrary focal elements. The results are not shown here as we can get similar conclusions
from the results:} The reliability estimation process of these methods
takes more time compared with that of $\LNS$ rule.
The behavior of these two methods is similar to that of $\LNS$ rule
when the parameter $k$ or $\lambda$ is set to be in a fixed range. But they are much more time-consuming compared with
$\LNS$ rule. {This confirms that the reliability discounting method in $\LNS$ rule is effective for the following conjunctive combination.}



\begin{center} \begin{figure}[!thbp] \centering
		\includegraphics[width=.8\linewidth]{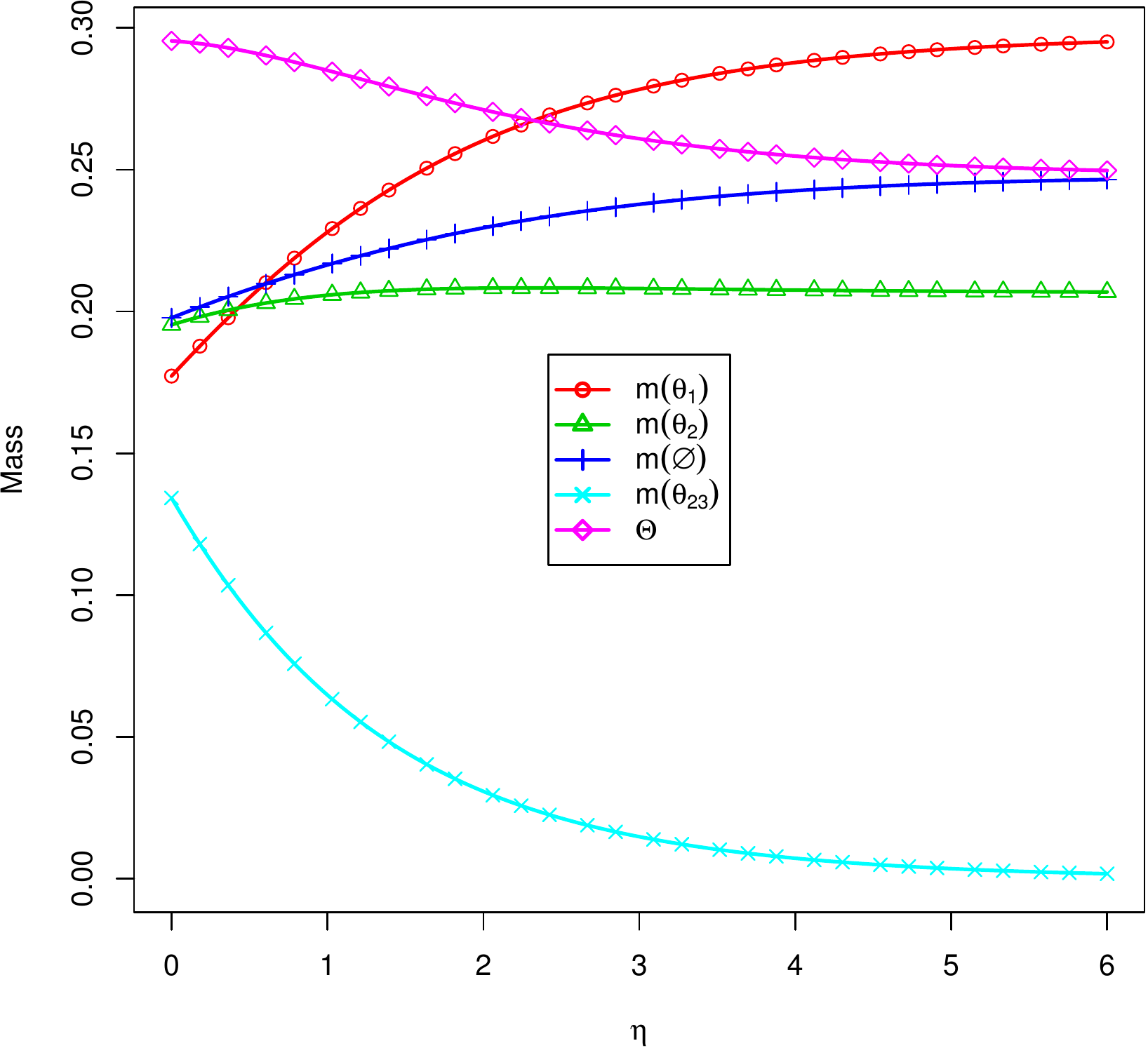}\hfill
       \parbox{1\linewidth}{\centering\small a. bba}

       \vspace{1em}
       \includegraphics[width=.8\linewidth]{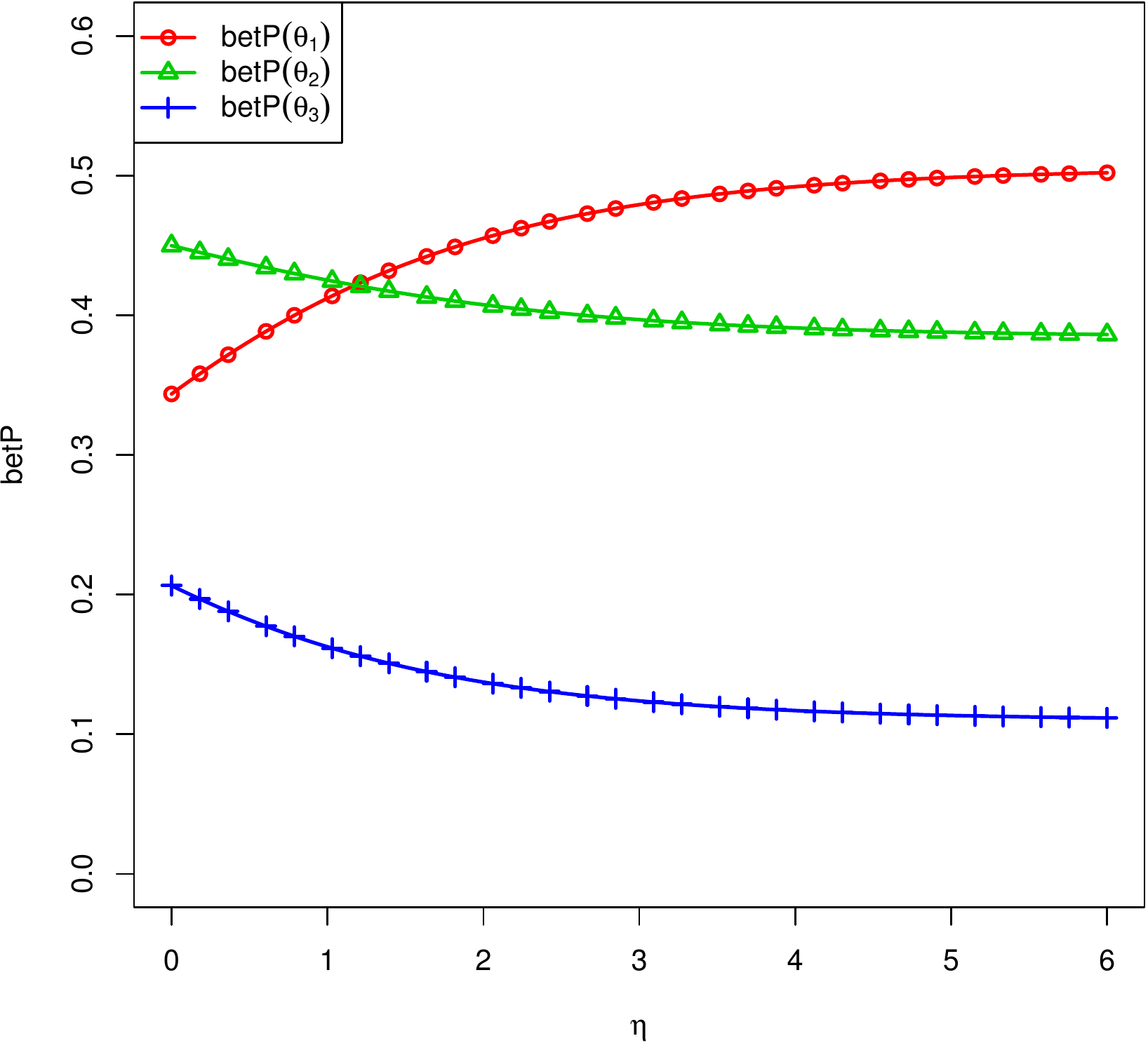} \hfill
       \parbox{1\linewidth}{\centering\small b. Pignistic  probability}
\caption{Combination results for three types of SSFs using $\LNS$ rule.  The mass functions
are generated randomly, and $\LNS$ rule is evoked with different values of $\eta$ ranging from 0 to 6.} \label{Exp5witheta} \end{figure} \end{center}

\Exp \change{~(The influence of parameter $\eta$).} We test here the influence of parameter  \change{$\eta$}  in {the} $\LNS$ rule. Simple support mass functions are
utilized in this experiment. Suppose that the discernment frame under consideration is $\Theta=\{\theta_1,\theta_2,\theta_3\}$. Three types
of SSFs are adopted. First $s_1 = 60$ and $s_2 = 50$ SSFs with focal elements $\{\theta_1\}$ and $\{\theta_2\}$ respectively (the other focal element is $\Theta$) are uniformly generated, and then
$s_3 = 50$ SSFs with focal element $\theta_{23}\triangleq  \{\theta_2, \theta_3\}$ are generated. The value of masses are randomly generated. Different values of $\eta$ {(see Eq.~\eqref{discountfactor})}
ranging from 0 to 6 are used to test. The
mass values in the fused bba by $\LNS$ varying with $\eta$ are displayed in Figure \ref{Exp5witheta}.a, and the corresponding pignistic probabilities
are shown in Figure \ref{Exp5witheta}.b.

\vspace{-2em}
From these figures, we can see that $\eta$ can have some effects on the final decision.  Figure \ref{Exp5witheta}.a shows that
with the increasing of {$\eta$}, \change{the mass assigned} to the singleton focal
elements increases. On the contrary, the mass given to the focal
element whose cardinality is bigger than one decreases.
In fact parameter {$\eta$} in $\LNS$ aims at  weakening the imprecise
evidence which gives only positive mass to focal elements with high cardinality, and
the exponent $\eta$ allows  to control the degree of discounting. If $\eta$ is larger,  {more
weight is given} to the  sources of evidence whose focal elements are more specific, and more discount will be committed to the imprecise
evidence. As a result, in the experiment when $\eta$ is larger
than 1.2, $\BetP(\theta_1) > \BetP(\theta_2)$ (Figure \ref{Exp5witheta}.b). At this time
the mass functions with focal element $\{\theta_2, \theta_3\}$ make little contribution
to the fusion process, while the final decision mainly depends on
the other two types of simple support mass functions with singletons as focal elements.

In real applications, $\eta$ could be determined based on  specific requirement. This work is not specially focusing on
how to determine $\eta$, thus in the following experiment we will set \linebreak $\eta=1$ as default.

\Exp \change{~(The principle for the global conflict).} 
The goal of this experiment is to show how Dempster's degree of conflict is dealt with by  most of rules when combining a large number of conflicting sources.

In this experiment, the frame of discernment is set to \linebreak
$\Theta = \{ \theta_1, \theta_2\}$. Assume that there are only 2 focal elements on each bba.
One is the whole frame $\Theta$, and the other is any of  the singletons ($\{\theta_1\}$ or $\{\theta_2\}$). The number of bbas which have the focal
element $\{\theta_1\}$ is {denoted by} $s_1$, while that with $\{\theta_2\}$ is $s_2$. We {first} fix
the value of $s_2$, and let $s_1 = t * s_2$, with $t$ a positive integer.
We generate $S = s_1 + s_2$ such kind of bbas randomly, but only withholding
the bbas for which the mass value assigned to $\{\theta_1\}$ or $\{\theta_2\}$ is greater than 0.5.

\begin{center} \begin{figure}[!thbp] \centering
		\includegraphics[width=0.8\linewidth]{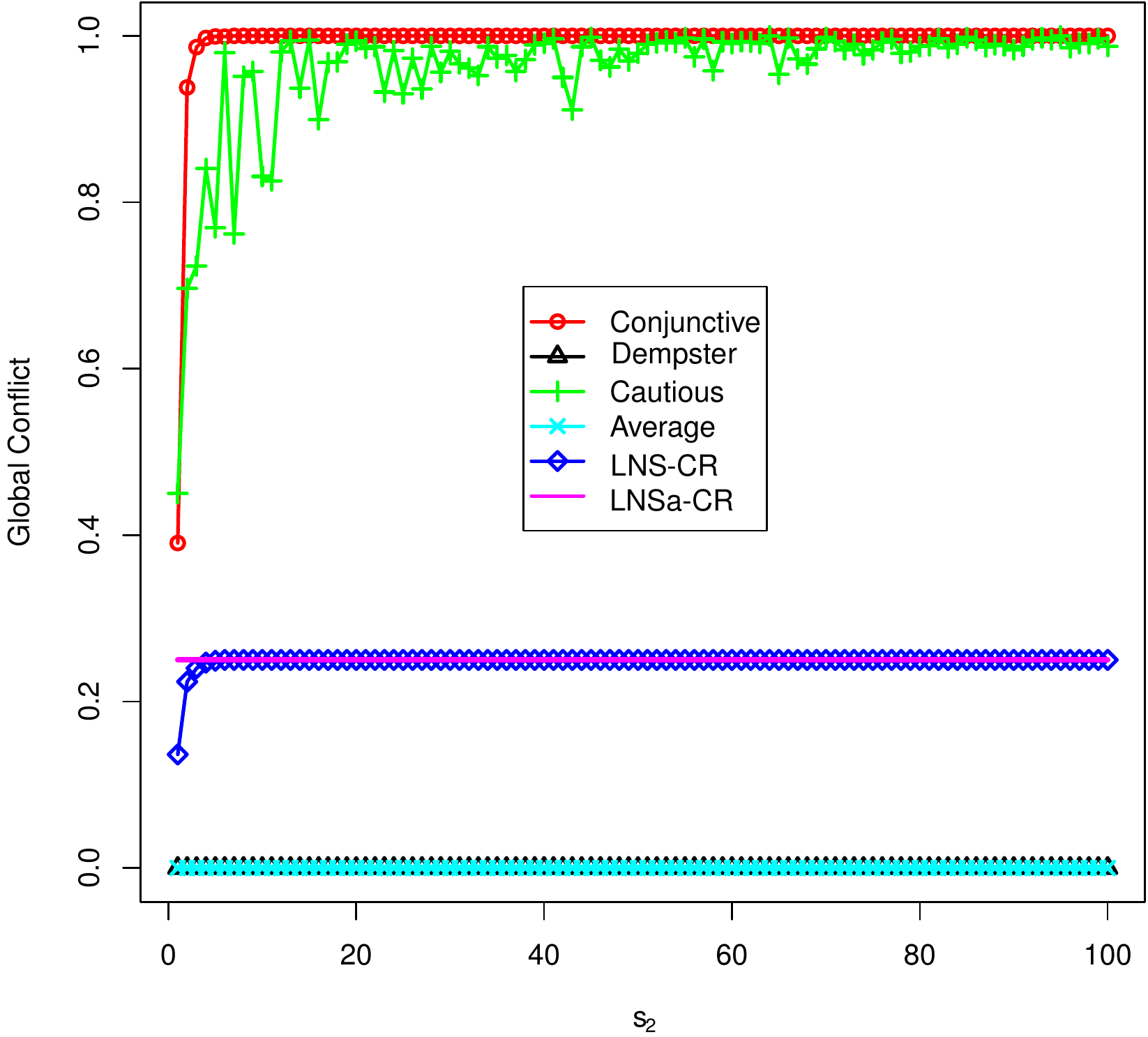}\hfill
\caption{{The global conflict after the combination with $s_2$ ranging from [0,100] and $s_1 = s_2$.}} \label{confwithn1} \end{figure} \end{center}

\begin{center} \begin{figure}[!thbp] \centering
		        \includegraphics[width=0.8\linewidth]{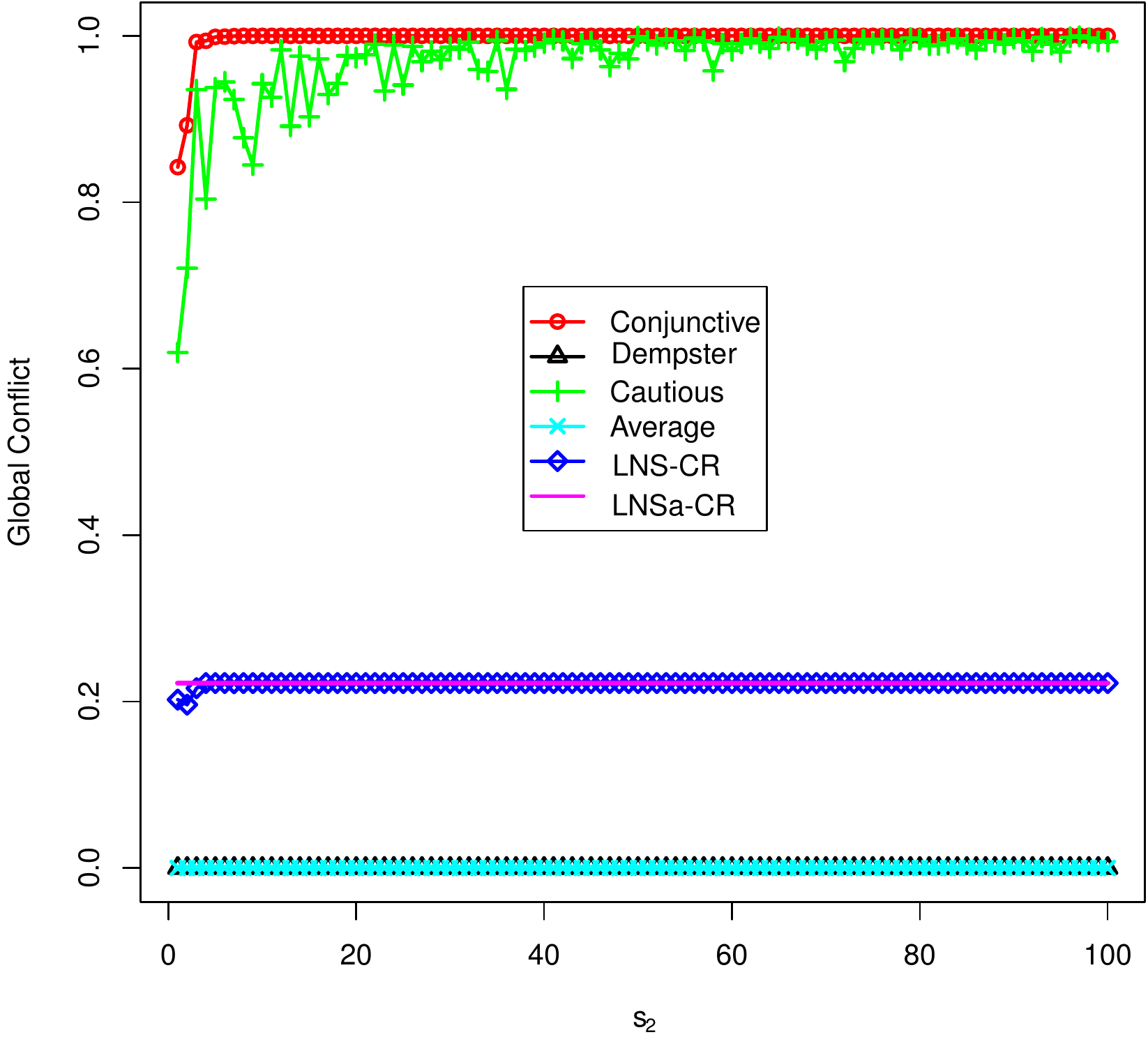} \hfill
\caption{{The global conflict after the combination with $s_2$ ranging from [0,100] and  $s_1 = 2*s_2$.}} \label{confwithn2} \end{figure} \end{center}

\begin{center} \begin{figure}[!thbp] \centering
	\includegraphics[width=0.8\linewidth]{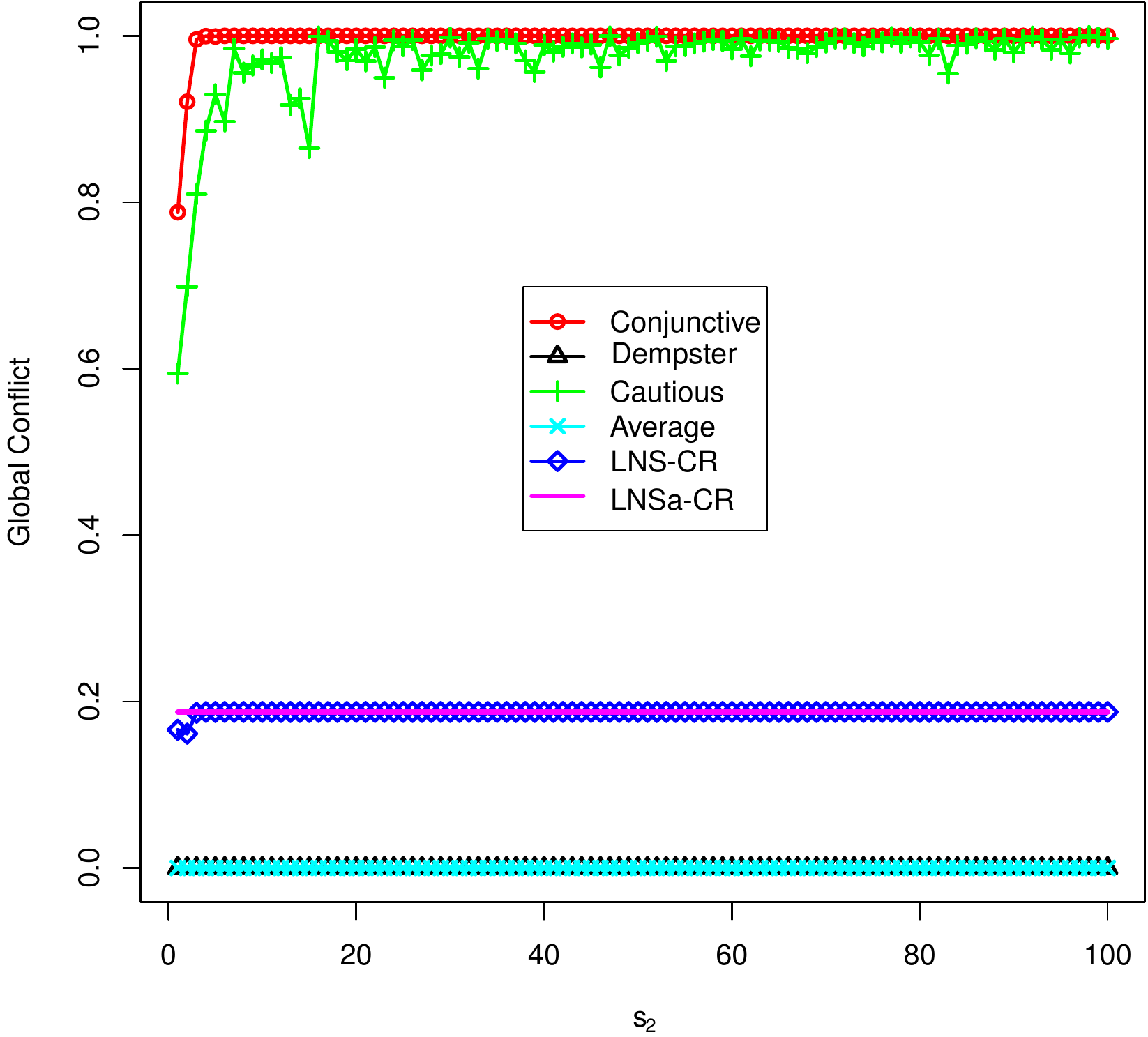}\hfill
	\caption{{The global conflict after the combination with $s_2$ ranging from [0,100] and  $s_1 = 3 * s_2$.}} \label{confwithn3} \end{figure} \end{center}

\begin{center} \begin{figure}[!thbp] \centering
        \includegraphics[width=0.8\linewidth]{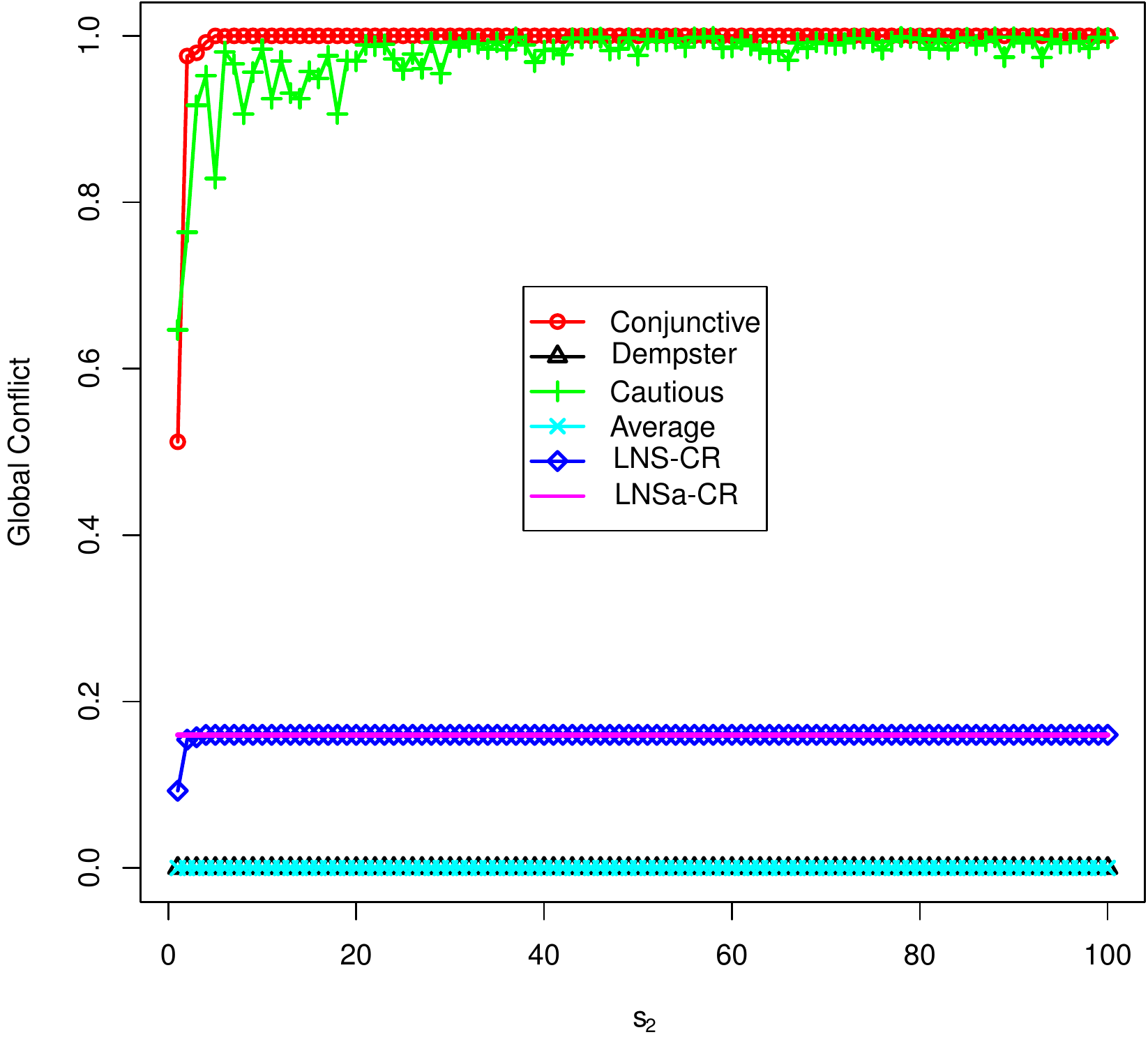} \hfill
\caption{{The global conflict after the combination with $s_2$ ranging from [0,100] and  $s_1 = 4 *s_2$.}} \label{confwithn4} \end{figure} \end{center}

\begin{center} \begin{figure}[!thbp] \centering
		\includegraphics[width=.8\linewidth]{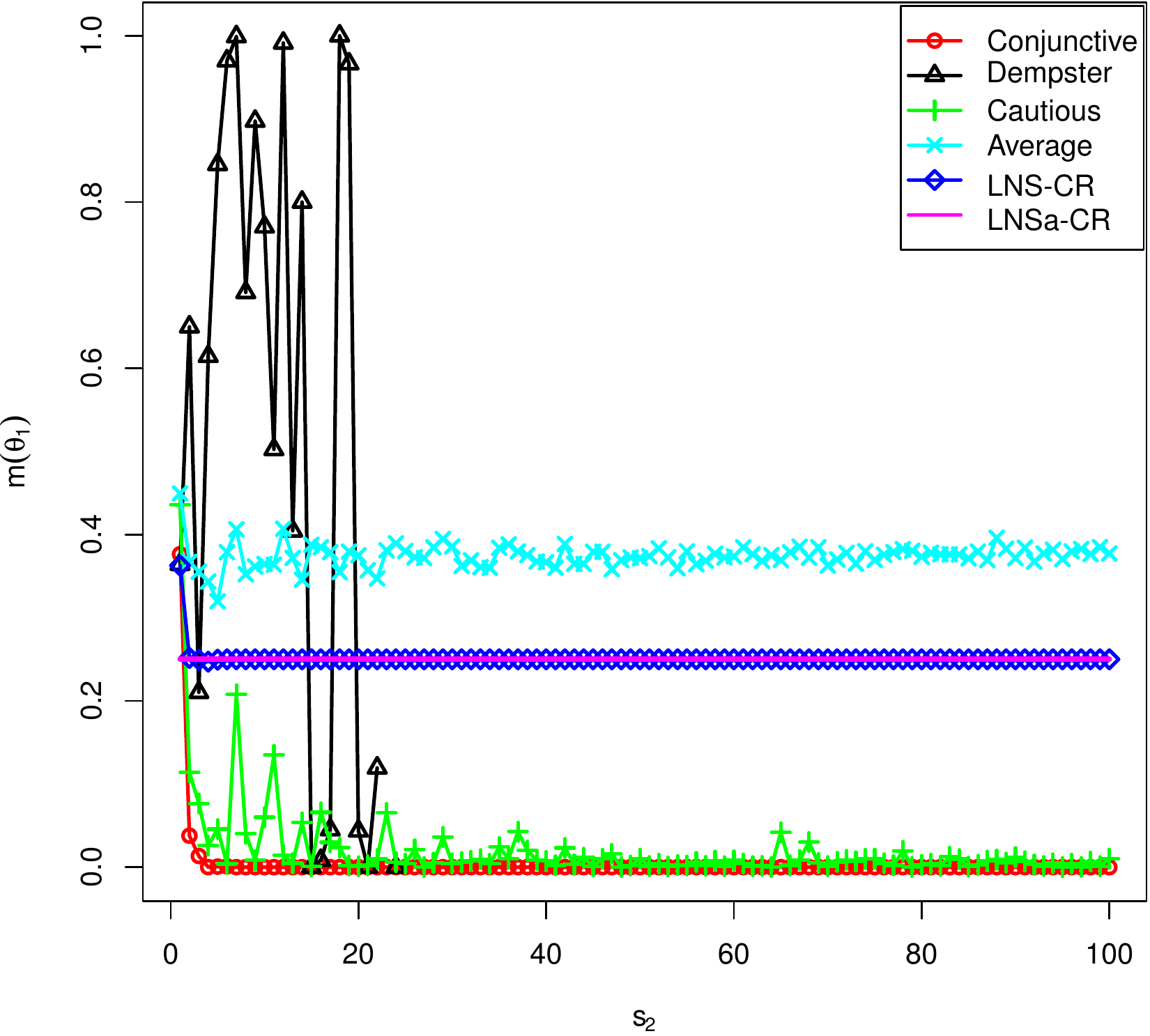}\hfill
\caption{{The mass on  $\{\theta_1\}$ after the combination with $s_2$ ranging from [0,100] and  $s_1 = s_2$.}} \label{theta1withn1} \end{figure} \end{center}

\begin{center} \begin{figure}[!thbp] \centering
		\includegraphics[width=.8\linewidth]{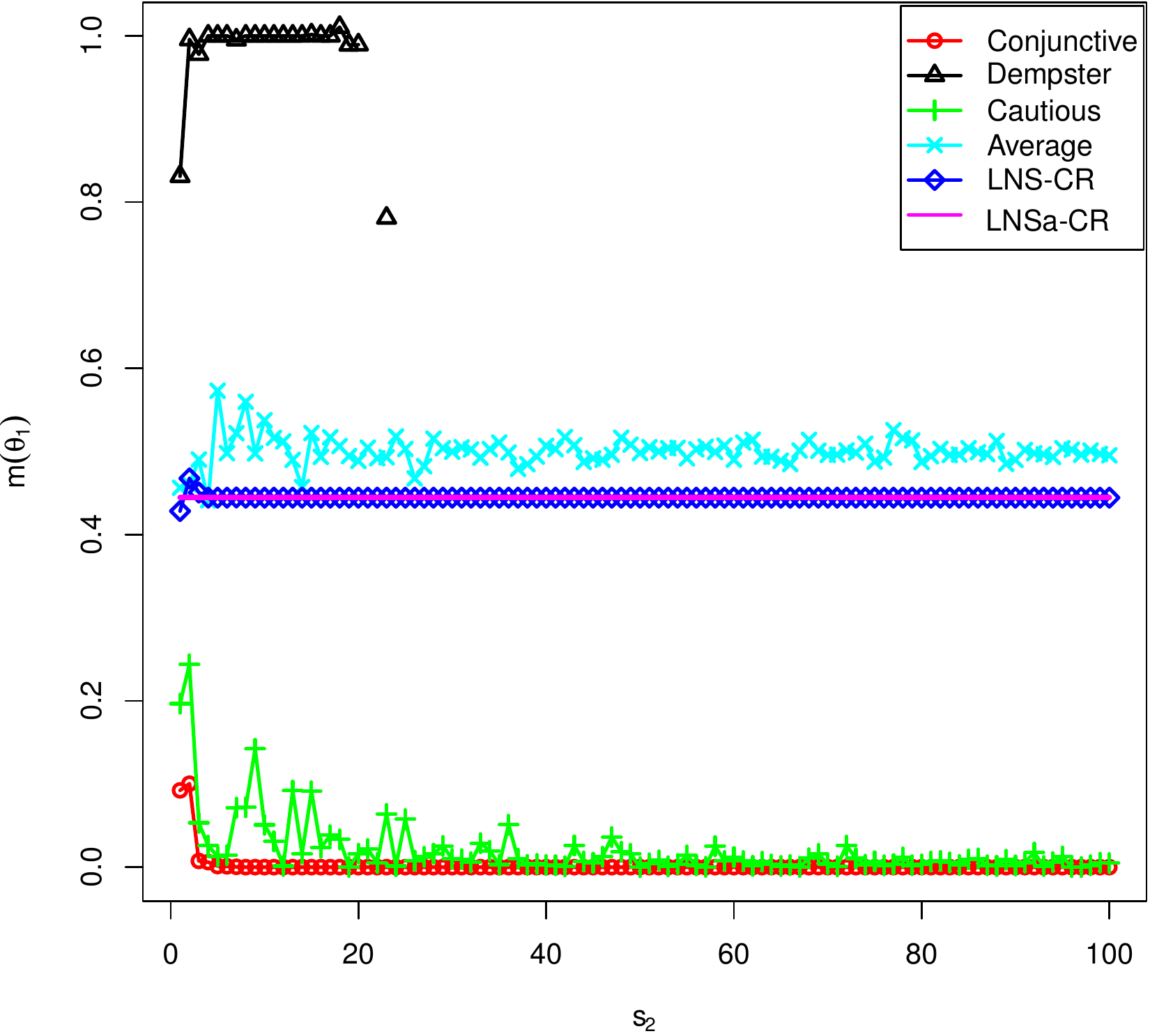} \hfill
\caption{{The mass on  $\{\theta_1\}$ after the combination with $s_2$ ranging from [0,100] and $s_1 = 2*s_2$.}} \label{theta1withn2} \end{figure} \end{center}

\begin{center} \begin{figure}[!thbp] \centering
	\includegraphics[width=.8\linewidth]{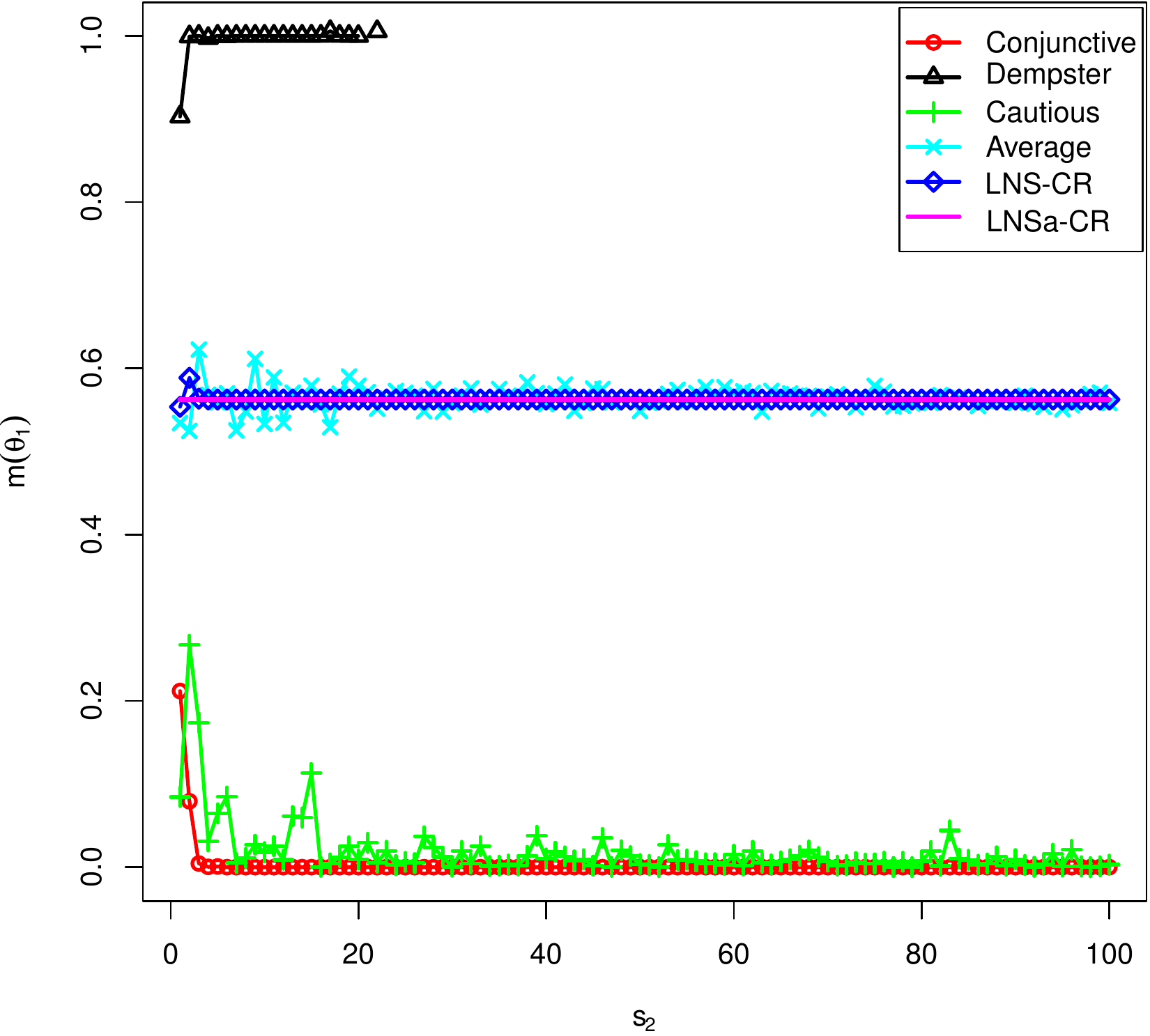}\hfill
\caption{{The mass on  $\{\theta_1\}$ after the combination with $s_2$ ranging from [0,100] and  $s_1 = 3 * s_2$.}} \label{theta1withn3} \end{figure} \end{center}

\begin{center} \begin{figure}[!thbp] \centering
	    \includegraphics[width=.8\linewidth]{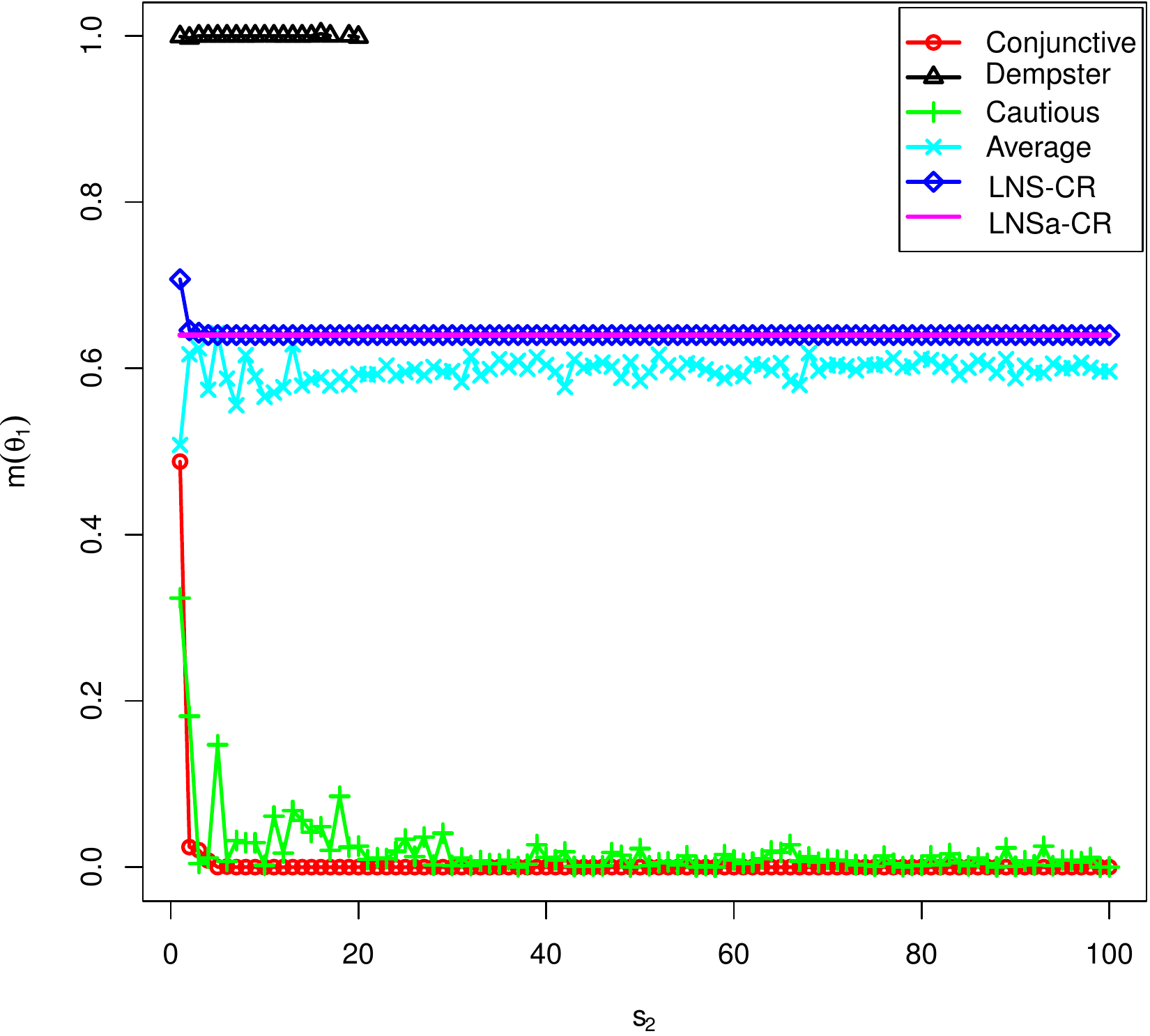} \hfill
\caption{{The mass on  $\{\theta_1\}$ after the combination with $s_2$ ranging from [0,100] and  $s_1 = 4 *s_2$.}} \label{theta1withn4} \end{figure} \end{center}

Four values of $t$ are considered here: $t=1, 2, 3, 4$. If $t=1$, $s_1 = s _2 = S/2$.  If $t=2$, the number of mass functions
supporting $\{\theta_1\}$ is two
times of that supporting $\{\theta_2\}$, and so on.   The global conflict (mass given to the empty set) after the combination with different values of $s_2$ for the four cases is displayed in Figures \ref{confwithn1}-- \ref{confwithn4} respectively. The mass assigned to the focal element $\{\theta_1\}$ with different combination approaches is shown in Figures \ref{theta1withn1} -- \ref{theta1withn4}.

\vspace{-2em}
It is intuitive that when $t$ becomes larger, the global conflict should be smaller and we should give more belief to the focal element $\{\theta_1\}$. From  Figures~\ref{confwithn1} -- \ref{theta1withn4} we can see that only the results by $\LNS$ rule are in accordance with this common sense.  The simple average rule assigns larger bba to $\{\theta_1\}$, but it does not keep any conflict. {In Figures \ref{theta1withn1} -- \ref{theta1withn4}, the mass given to $\{\theta_1\}$ by $\DS$ rule cannot be displayed when $S$ is large (and also for some small $S$), because in these cases the global conflict is 1 and the normalization could not be processed.  As we can see, $\DS$ rule could not work at all when $s_2$ is larger than 20.}  Although the conjunctive rule and cautious rule could work when combining a larger number of mass functions,  the
obtained fused mass function is $m(\emptyset)\approx 1$, which is useless for {decision in practical situations.}

{The results also confirm the  equivalent of the $\LNS$ rule and $\LNSA$ rule when the number of sources is large, although the
results provided by the two rules are not the same when there are not many mass functions to combine. }
From Figures \ref{confwithn1} -- \ref{confwithn4}
we can see a kind of limit of the global conflict for the $\LNS$ rule. In fact, the mass on the empty set for this
rule {depends on} the size of the frame of discernment and more directly on the number of groups created in the first step of the rule.
The limit value of the global conflict will tend to 1 with the increase of the size of discernment when considering only categorical bbas on
different singletons.

\begin{center} \begin{figure}[!thbt] \centering
 		\includegraphics[width=0.8\linewidth]{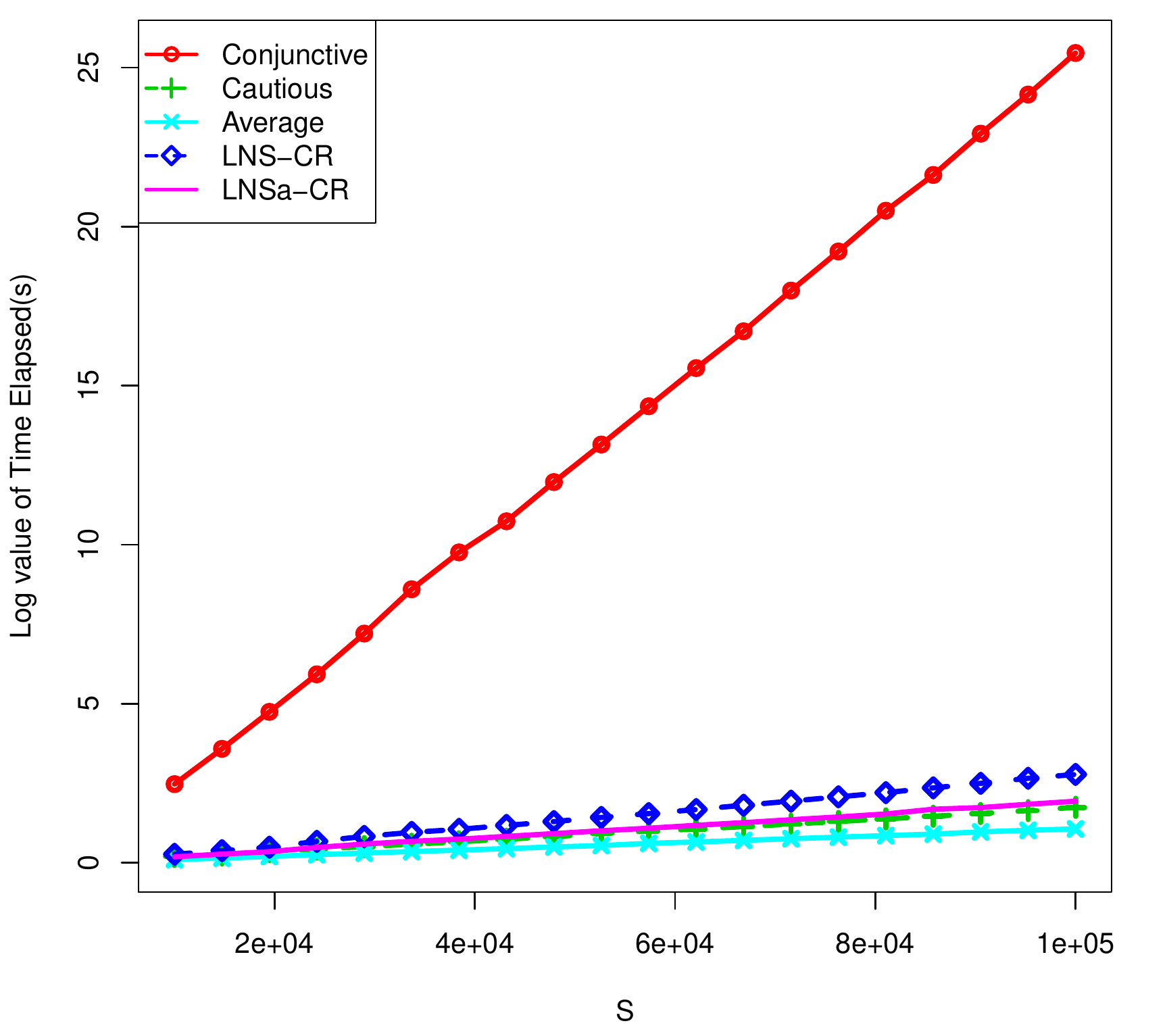}\hfill
         \parbox{1\linewidth}{\centering\small a. Time lapse by five different rules} \hfill

         \includegraphics[width=0.8\linewidth]{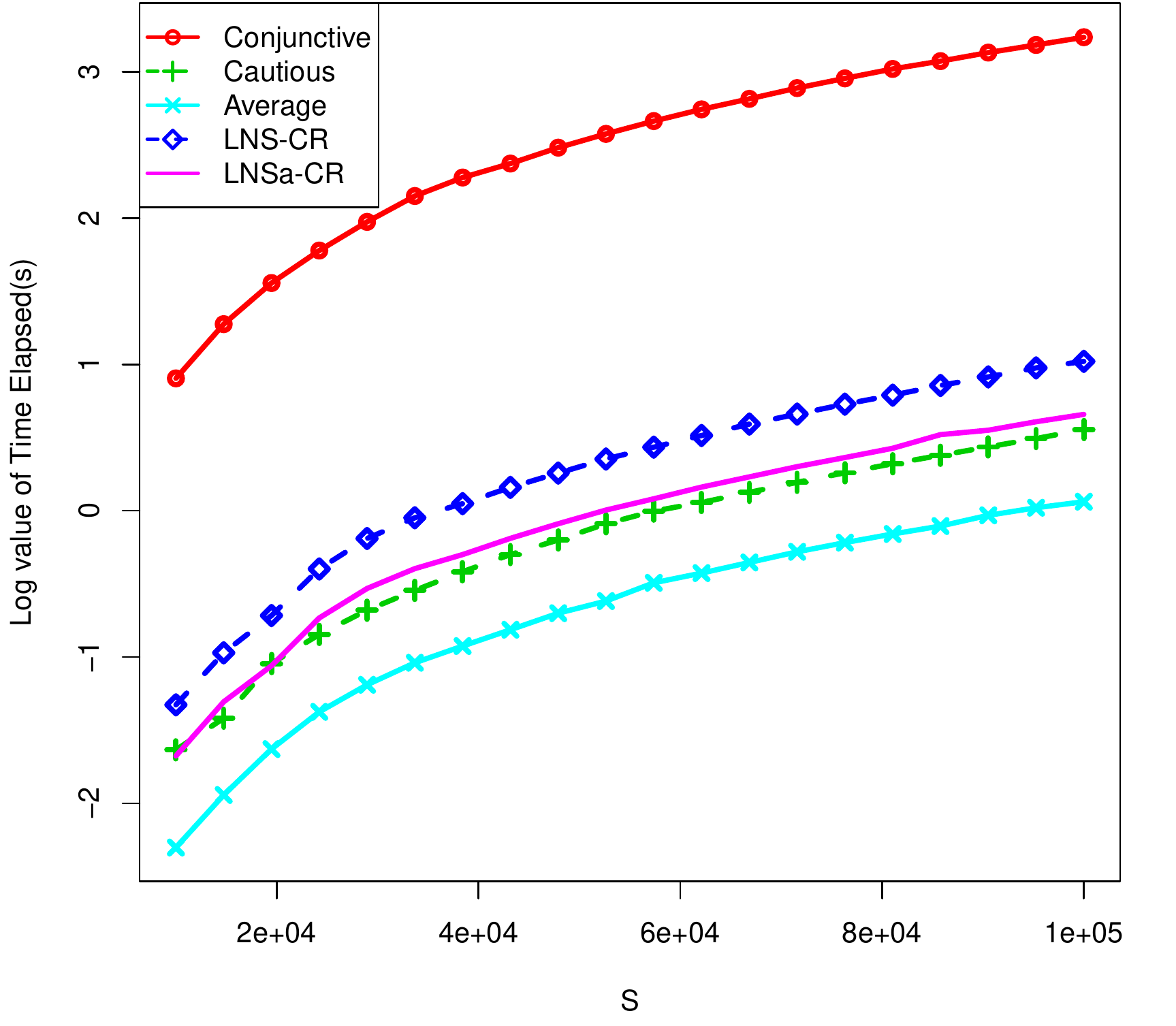}\hfill
           \parbox{1\linewidth}{\centering\small ~~~~~~b. The $\log$ value of Time lapse  by five different rules}
 \caption{{Time lapse for combining  SSFs.}}
 \label{timecomSSF} \end{figure} \end{center}

\Exp \change{~(The complexity).} 
In this experiment, the complexity of $\LNS$ rule will be compared with other combination rules in terms of time consumption.
Simple support mass functions defined on a frame of discernment with eight elements
are considered first.  The focal elements of each bba are set
to be a random subset of $\Theta$ and  $\Theta$ itself. The time elapsed (and also the $\log$ value of the time elapsed) with the
number of sources $S$ varying from 10,000 to 100,000 is shown in Figure \ref{timecomSSF}\footnote{The result of Dempster rule is the
same as that of conjunctive rule.}. We can see that the running time of $\LNS$ is much smaller than that of the conjunctive rule. $\LNSA$ rule
takes almost the same time as cautious rule.
Average rule is the best among the five rules.
As $S$ increases, the application of $\LNSA$ rule can save more time compared with the use of $\LNS$ rule.
The increment of time consumption  with respect to $S$ is moderate.
This tends to show that $\LNS$ rule is suitable for
combining a large number of SSFs. Remark that the decomposition process is not required
when the cautious rule or $\LNS$(a) rule  is adopted for combining SSFs.

\begin{center} \begin{figure}[!thbt] \centering
 		\includegraphics[width=.8\linewidth]{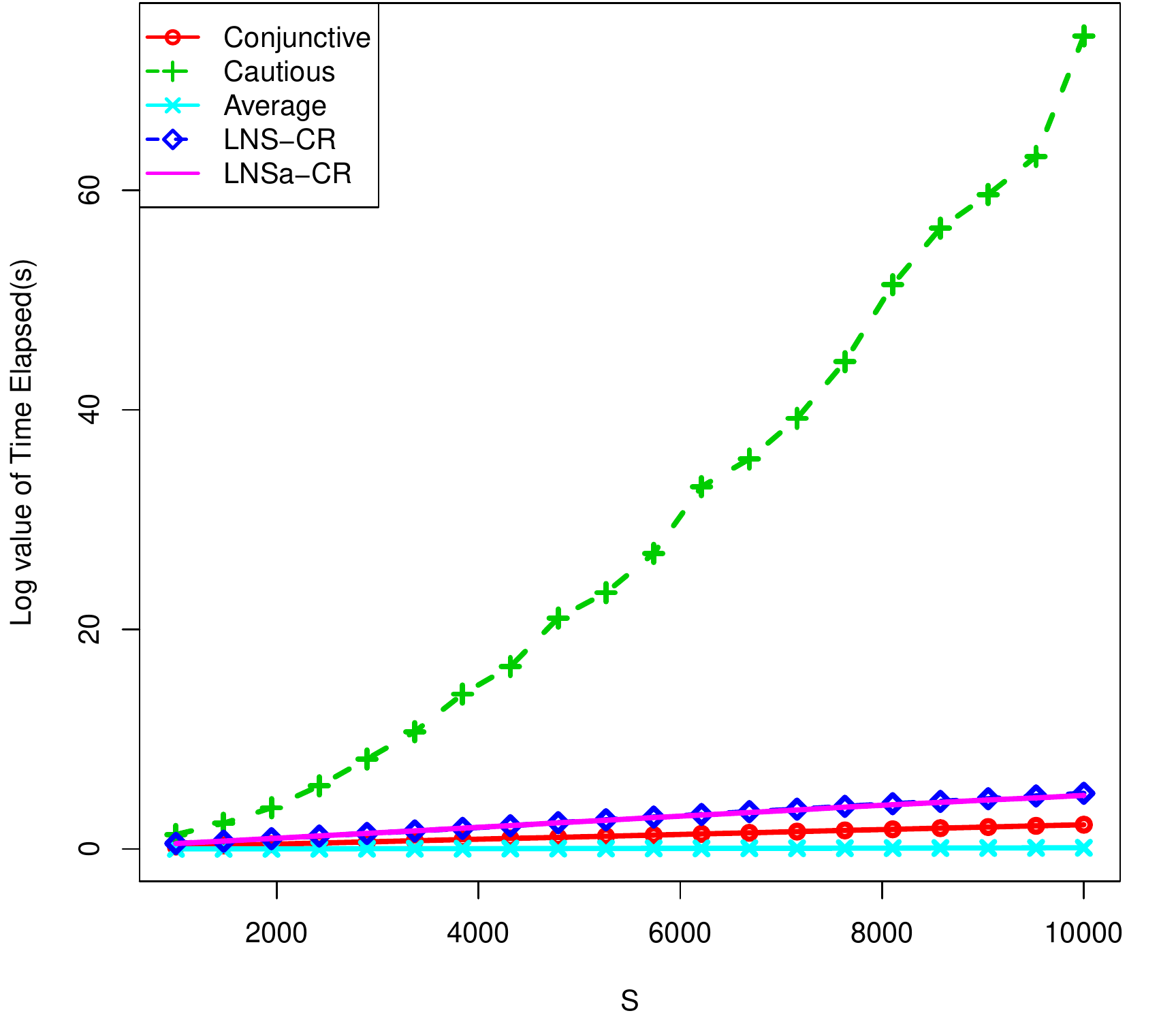}\hfill
   \parbox{1\linewidth}{\centering\small ~~~~~~~a. Time lapse by five different rules} \hfill

 \includegraphics[width=.8\linewidth]{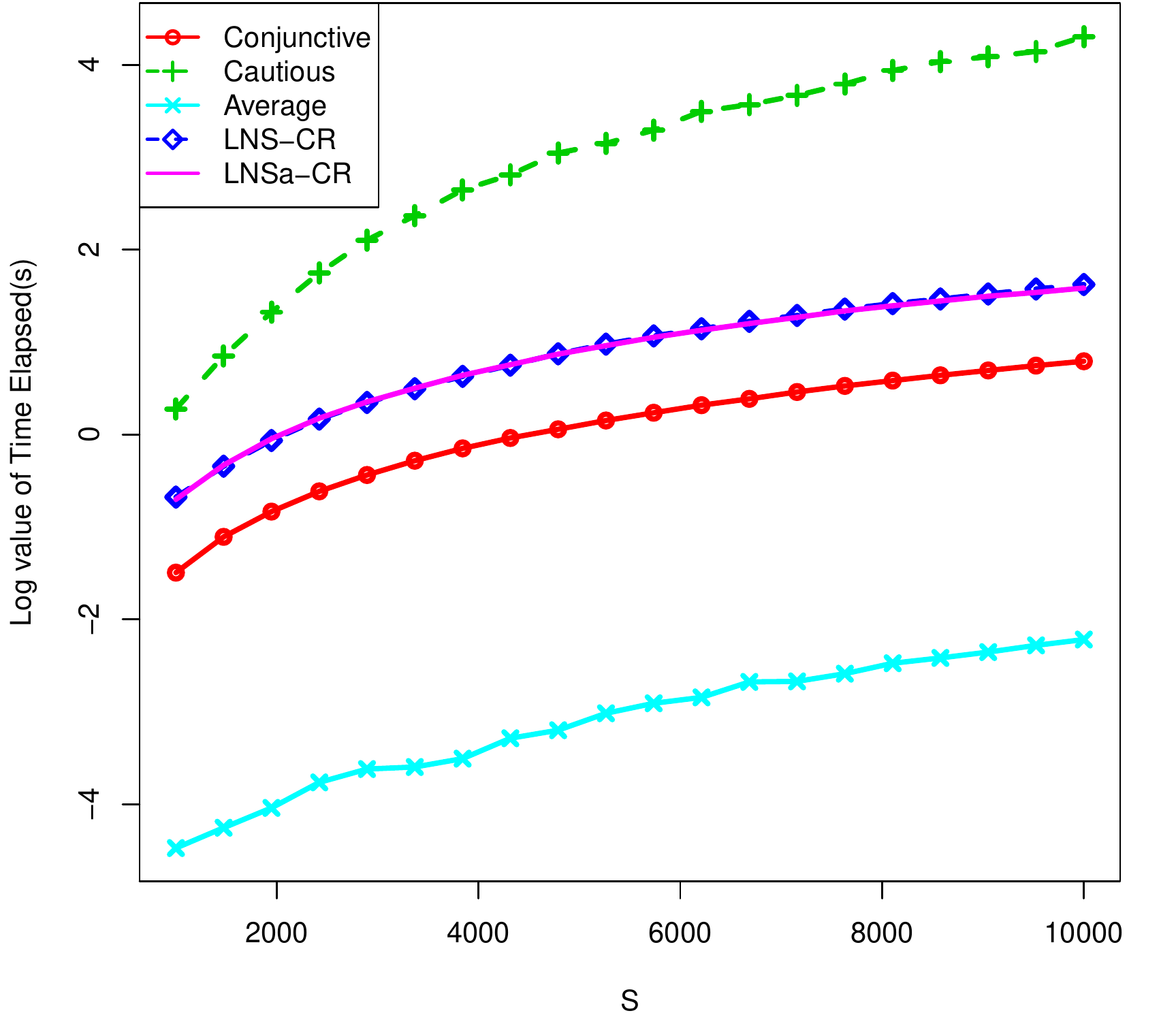}\hfill
          \parbox{\linewidth}{\centering\small~~~~~b. The $\log$ value of Time lapse by five different rules}
 \caption{{Time lapse for combining consonant bbas.}} \label{timecomExp4} \end{figure} \end{center}

\begin{center} \begin{figure}[!thbt] \centering
 		\includegraphics[width=0.8\linewidth]{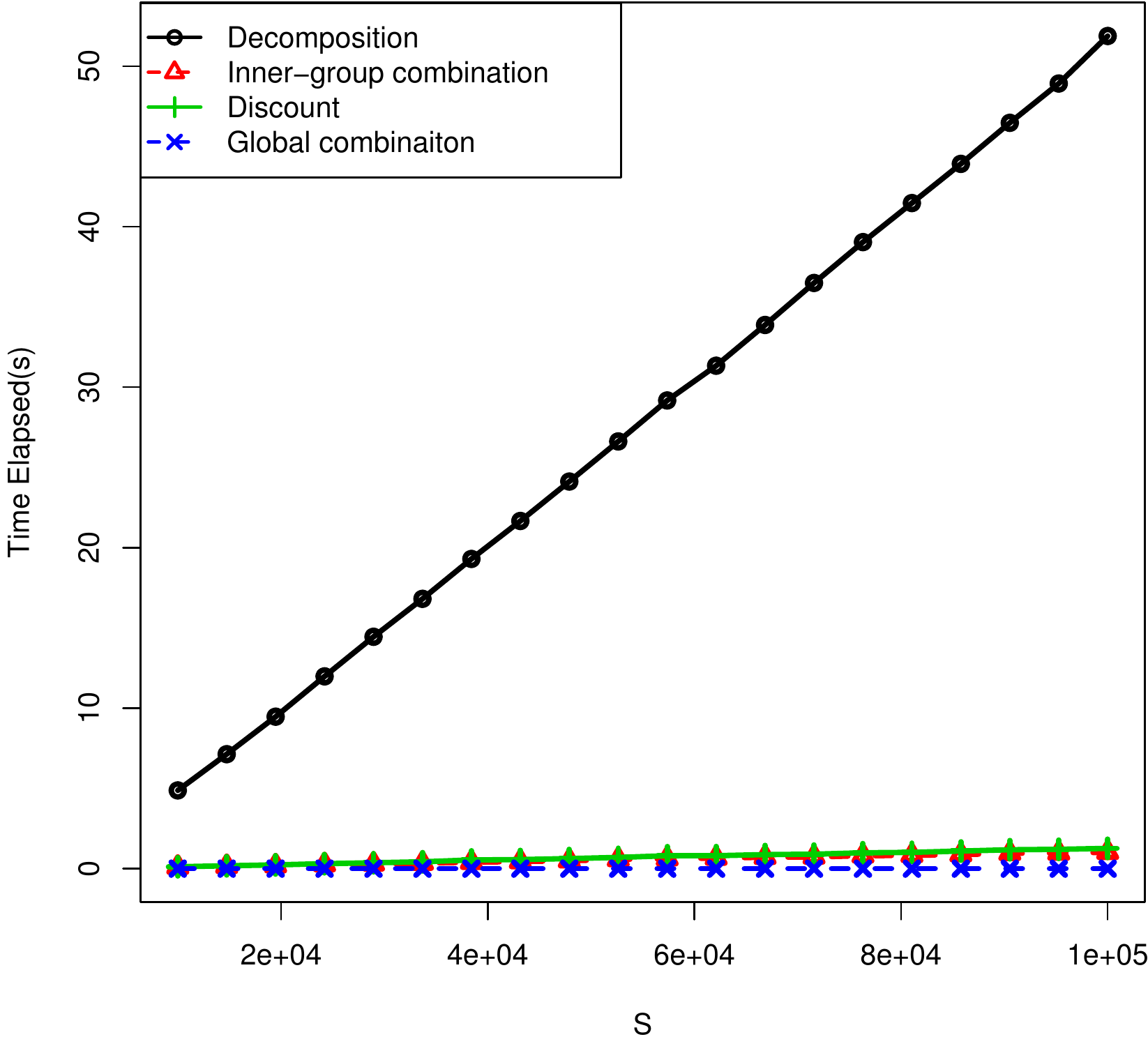}\hfill
 \caption{Time lapse of each step using LNS combination rule
 with $S$ varying from 10,000 to 100,000.} \label{timecomStep} \end{figure} \end{center}

\vspace{-2em}
As mentioned before, for the combination of general separable mass functions (not SSFs), $\LNS$ needs four steps: decomposition, inner-group combination, discounting and global combination. The difference between  the combination of
any kind of separable bbas and  of SSFs is the decomposition process, which is not necessary for the latter.
We have designed another experiment on consonant bbas\footnote{All consonant bbas are separable.} over a frame of discernment
with eight elements, and the number of focal elements is set to 5. The  focal elements are randomly set to five nested
subsets of $\Theta$, and the mass
values are generated uniformly. The average running time (and the $\log$ value of the running time) of 10 trials by the use of different combination rules with different number of sources $S$
is displayed in Figure
\ref{timecomExp4}.a (and Figure
\ref{timecomExp4}.b)\footnote{The result of cautious rule is not displayed for large $S$, as it
has been already shown that cautious rule is significantly worse than the other rules in terms of time consumption when $S$ is small.}. In order to
show the complexity of $\LNS$ rule more clearly, the elapsed time  in each of the four steps is shown in Figure \ref{timecomStep}.


\vspace{-4em}
As we can see from these figures, the time consumption
 of $\LNS$ is significantly smaller than the cautious rule, but a little worse than the conjunctive rule
and the average rule.  Although the complexity of cautious rule is the same as $\LNS$ rule and both of them require
a decomposition process, it takes more running time than $\LNS$ rule. The reason may be the different
combination approach for the mass functions in the same group. The complexity of that process by cautious rule is $O(S2^n)$ (The calculation
is to find the minimum of each row in a $S\times 2^n$ matrix), while for $\LNS$
is $O(S)$. $\LNSA$ is faster
than $\LNS$ when $S$ is large. Figure \ref{timecomStep} shows that the most time-consuming step in $\LNS$ rule is the decomposition.
Moreover as $S$ increases, the
increase of time lapse for the inner-group combination, discount, and global combination is limited. This is compliant with
the complexity analysis of each step for $\LNS$ rule in Section \ref{secproperties}. In many applications
the mass functions are directly SSFs in which case
there is no need to perform the decomposition, and $\LNS$ is the best choice to fuse a large number of bbas.

{
\section{Perspective on applications}
Pattern recognition is a class of problems where the theory of belief functions  has proved to allow
increased performances \cite{denoeux1995k}.
In such  problems we can be facing  many bbas to combine. \citet{denoeux1995k} proposed Evidential KNN method (EKNN)  as an extension of KNN in the framework of the theory of belief functions to better model the uncertainty in neighbor point interactions. The $\DS$ rule is adopted to combine the mass evidence from $K$ neighbors in EKNN.}

{
The problem considered here is to  classify an input pattern $\x$ into $n$ categories or classes, denoted by $\Theta=\{\theta_1,\theta_2,\cdots,\theta_n\}$. The available information is assumed to consist of a training set
$\mathcal{L} = \left\{(\x^{(1)}, \theta^{(1)}),(\x^{(2)}, \theta^{(2)}),\cdots,(\x^{(N)}, \theta^{(N)})\right\}$ of $N$ patterns $\x^{(i)}$
$i=1,2,\cdots,N$ with known class labels $\theta^{(i)} \in \Theta$. To classify pattern $\x$, each pair $(\x^{(i)}, \theta^{(i)})$  constitutes
a distinct item of evidence regarding the class membership of $\x$. If the $K$ nearest neighbors according to the distance measure are considered,
$K$ items of evidence  can be obtained. These bbas can be constructed according to a
relevant metric between pattern $\x$ and its $j^{\text{th}}$ neighbor $\x^{(i)}$
\begin{align}
\label{eknnbba}
  & m_i(\{\theta_q\}) = \alpha \phi(d^{(i)}), \nonumber \\
  & m_i(\Theta) = 1 - \alpha \phi(d^{(i)}), \nonumber \\
  & m_i(A) = 0~~ \forall A \in 2^\Theta \setminus \{\{\theta_q\},\Theta\},
\end{align}
where $d^{(i)}$ is the (Euclidean) distance between $\x$ and its $j^\text{th}$ neighbor $\x^{(i)}$ with class label $\theta^{(i)}=\theta_q$, $\alpha$
 is a discounting parameter and $\phi(\cdot)$ is a decreasing function on $\mathbb{R}^+$ defined as
\begin{equation}
  \phi(d^{(i)}) = \exp \left(-\gamma_q \left(d^{(i)}\right)^2\right)
\end{equation}
with $\gamma_q$ being a positive parameter associated to class $\theta_q$. It
can be heuristically set to the inverse of the mean Euclidean
distance between training data belonging to class $\theta_q$.  In
EKNN, the $K$ bbas for each neighbor are aggregated using the $\DS$ rule
to form a resulting bba. A
decision has to be made regarding the assignment of sample
$\x$ to one individual class.  The
maximum of pignistic probability can be used for decision-making.
}
{
\subsection{A small data set with noisy training sample}
Figure \ref{circle_example_ori} illustrates a simple two-class (red circle and green triangle) data set, where  there are seven objects in each class. The pattern $\x$ marked by blue star  is the sample data to be classified. The $K$ bbas using the distance to its neighbor could be constructed by Eq.~\eqref{eknnbba}, and the five nearest neighbors  are denoted by $N_i$ orderly in the figure. Set $\alpha = 0.95$ and $\gamma_i$ is the inverse of the average distance between the points in class $\theta_i$, $i=1,2$. The fused mass function by different combination rules with $K=4$ and $K=5$ are listed in Table \ref{classifierexamplemassk=4}  and \ref{classifierexamplemassk=5} respectively.
\begin{center} \begin{figure}[!thbt] \centering
		\includegraphics[width=0.8\linewidth]{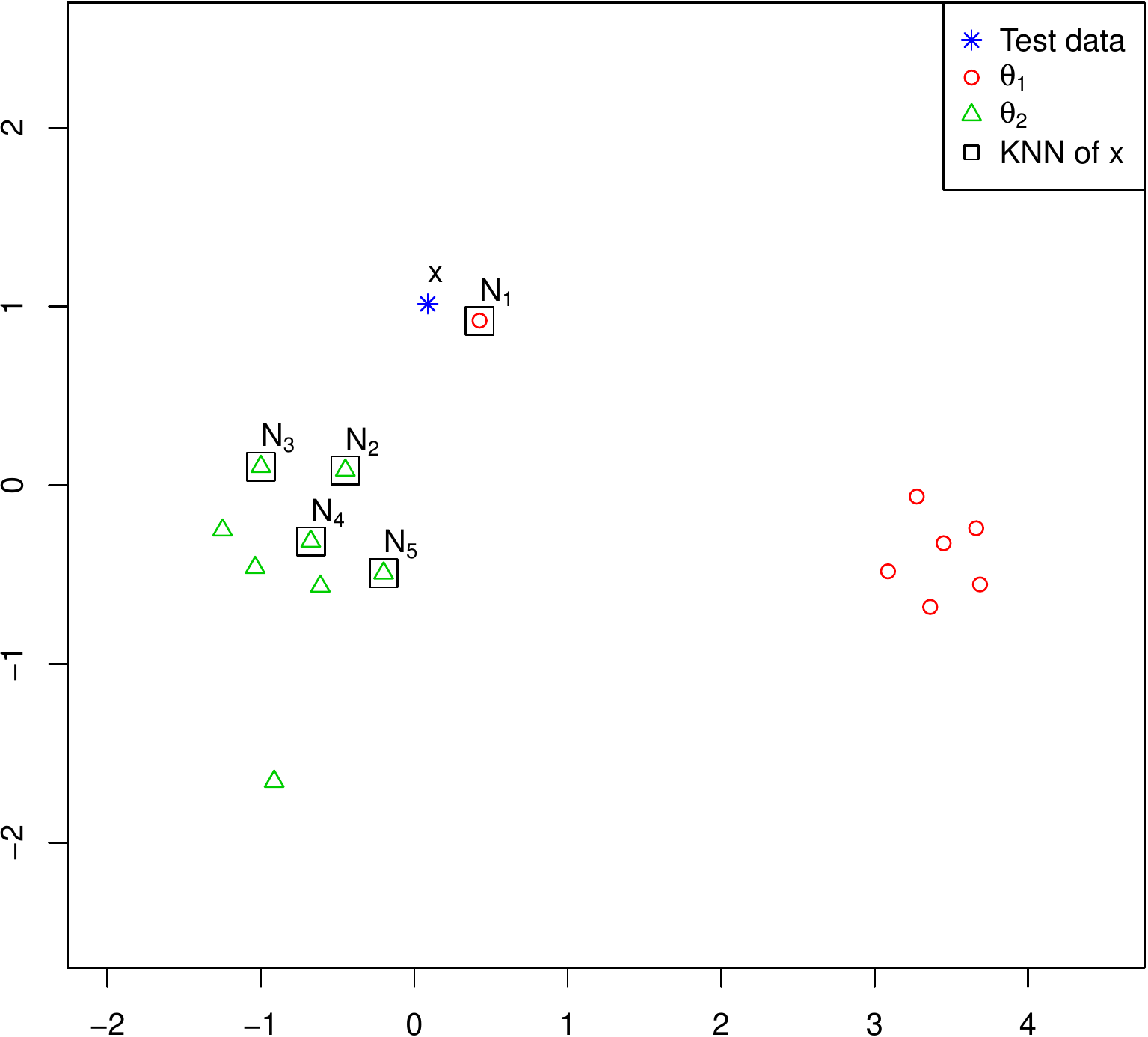}\hfill
\caption{A small data set.} \label{circle_example_ori} \end{figure} \end{center}
\begin{center} \begin{figure}[!thbt] \centering	
        \includegraphics[width=0.8\linewidth]{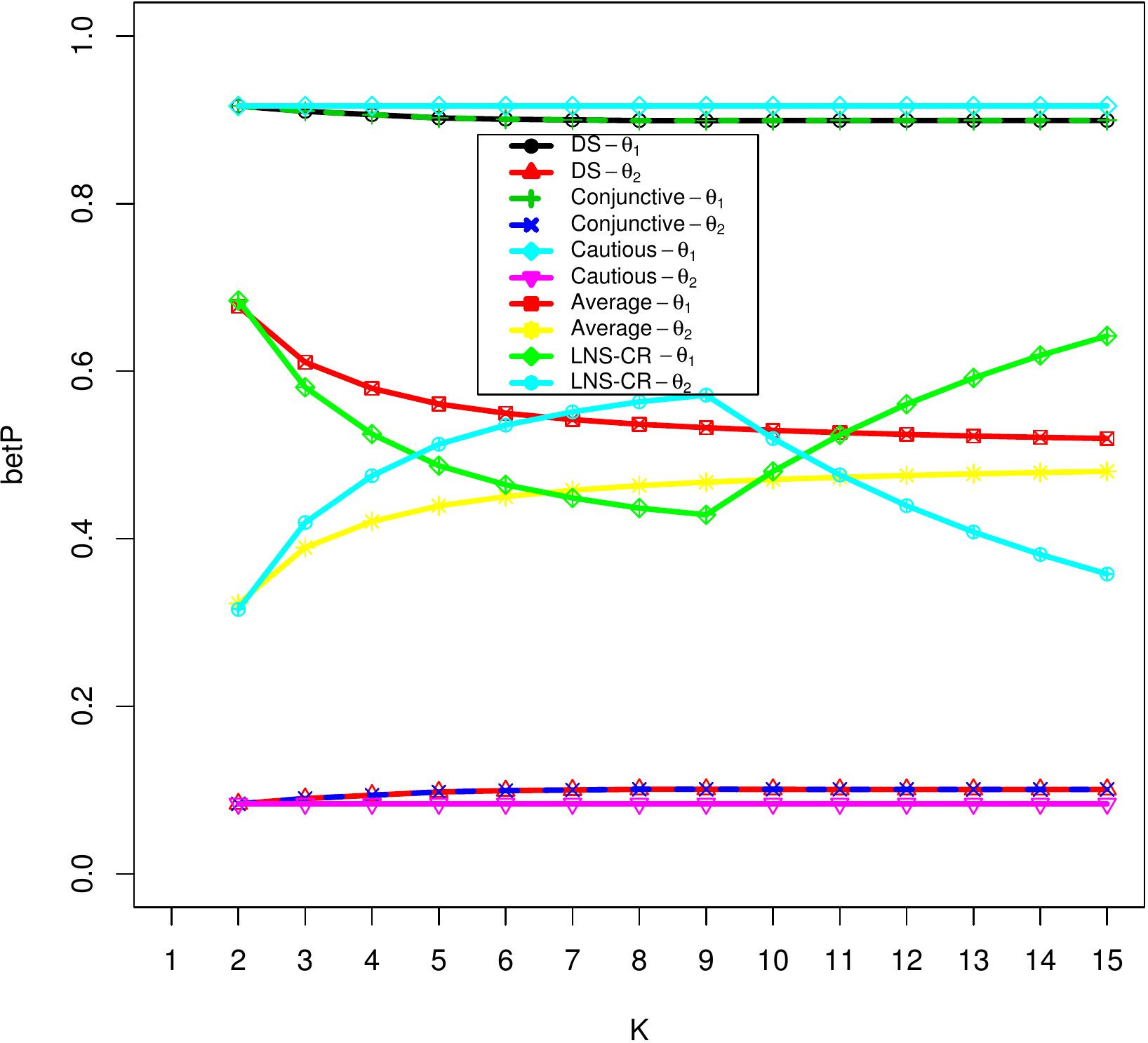}\hfill
\caption{Pignistic probability.} \label{circle_example_betp} \end{figure} \end{center}
}

{
As we can see from Figure \ref{circle_example_ori}, pattern $\x$ is closer to class $\theta_2$. Among pattern $\x$'s
five nearest neighbor $N_j, j=1,2,\cdots,5$, four belong to class $\theta_2$ while only
1 to class $\theta_1$. \change{The real class of object $N_1$ is  $\theta_1$}, but it is located in the boundary of the class and far from the other
data points in the class. It may be a noisy item of $\theta_1$. The standard
KNN rule can correctly classify object $x$ \change{to $\theta_2$} when $K>3$. However, if the evidential KNN model is applied,
due to the existence of a such neighbor, the behavior of the combination rules has been
affected. From Table \ref{classifierexamplemassk=4} we can see, when $K=4$, the fused
bbas by all combination rules all assign more  mass to $\theta_1$ than to $\theta_2$. Consequently,
pattern $\x$ will be classified into class $\theta_1$ if  the pignistic probability is considered for making
decision. The same phenomenon also occurs when $K$ is smaller than 4 (see Figure \ref{circle_example_betp}).
When $K=5$ (Table \ref{classifierexamplemassk=5}), only the $\LNS$ rule could partition pattern $\x$
into class $\theta_2$, which seems more reasonable.  The pignistic probabilities (Figure \ref{circle_example_betp}) by the $\DS$, conjunctive,
cautious and average rules for class $\theta_1$ are significantly higher  than those for class $\theta_2$, even when $K$ is large. These rules are
not robust to the noisy training data. Pattern $\x$ could be correctly classified to $\theta_2$ by $\LNS$ rule when $K$ is between 5 and 10.
}

{
It is indicated that when there are some noisy data in the training data set, the performance of the
combination rule may become worse with small $K$. We should increase $K$ moderately to improve the performance
of the classifier. But as we analyzed before, the existing
combination rules do not work well for aggregating a large number of mass functions. This is a limit of the use of evidential classifier.
}

{
\begin{table}[ht]
\centering \caption{The fused bba by different combination rules $(K=4)$.}
\begin{tabular}{lrrrrrr}
  \hline
 & Conjunctive & $\DS$ & Cautious & Average & $\LNS$ \\
  \hline
$\emptyset$ & 0.2009 & 0.0000 & 0.1473 & 0.0000 & 0.0377 \\
  $\{\theta_1\}$ & 0.6771 & 0.8473 & 0.7307 & 0.2195 & 0.1818 \\
  $\{\theta_2\}$ & 0.0279 & 0.0349 & 0.0205 & 0.0606 & 0.1339 \\
  $\Theta$ & 0.0941 & 0.1177 & 0.1015 & 0.7199 & 0.6466 \\ 
   \hline
\end{tabular} \label{classifierexamplemassk=4}
\end{table}
\begin{table}[ht]
\centering \caption{The fused bba by different combination rules $(K=5)$.}
\begin{tabular}{lrrrrrr}
  \hline
 & Conjunctive & $\DS$ & Cautious & Average & $\LNS$ \\
  \hline
$\emptyset$ & 0.2198 & 0.0000 & 0.1473 & 0.0000 & 0.0352 \\
  $\{\theta_1\}$ & 0.6582 & 0.8436 & 0.7307 & 0.1756 & 0.1404 \\
  $\{\theta_2\}$ & 0.0305 & 0.0391 & 0.0205 & 0.0541 & 0.1651 \\
  $\Theta$ & 0.0915 & 0.1172 & 0.1015 & 0.7703 & 0.6593 \\ 
   \hline
\end{tabular} \label{classifierexamplemassk=5}
\end{table}
}

{
\subsection{Real data sets}
In this section, we consider some well \change{known} real data sets from the UCI
repository\change{\footnote{\change{http://archive.ics.uci.edu/ml/datasets.html}}} summarized in Table \ref{ucidatalist}.
The classification rates by using different combination rules in evidential KNN model  are displayed in Figure \ref{Exp5classificationrateUCI}.}
\change{Note that the ``leave-one-out" method is adopted here to test the classifier.}

\begin{table}[ht]
\centering\caption{A summary of  UCI data sets.}
\begin{tabular}{lllllll}
  \hline
  Data set & No. of objects & No. of cluster & No. of attributes\\
  \hline
Iris & 150 & 3 & 4\\
Yeast  & 1484 & 10 & 8\\
Digits  & 5620 & 10 & 64\\
 \hline
\end{tabular}\label{ucidatalist}
\end{table}


{
As we can see from Figure \ref{Exp5classificationrateUCI}, for all the three data sets, the performance is almost the same for
the two combination rules, \change{$\LNS$ and DS}, in terms of classification rates. But there  is a little improvement by the use of $\LNS$ rule
when $K$ is large. To make it clear, we specially depict  the results on Digits data set in Figure \ref{Exp5classificationrateDigits}. It
is  shown that when $K>12$, the classification rates by the use $\LNS$ rule are a little
larger than those through DS rule. We show the mass given to the empty set (global conflict) after the combination using conjunctive
rule and $\LNS$ rule with different
values of $K$ in Figure \ref{digits_conflict}.  The $y$-axis is the maximal assignment to $\emptyset$ among all the mass
functions for the test data.  \change{As we can see, the global conflict tends to 1 quickly as $K$ increases, while $\LNS$ rule keeps a
moderate degree of global conflict.} As DS rule is a normalized conjunctive rule, there is not sense to normalize a mass assignment
with  high global conflict.
\begin{center} \begin{figure}[!thbt] \centering
		\includegraphics[width=0.8\linewidth]{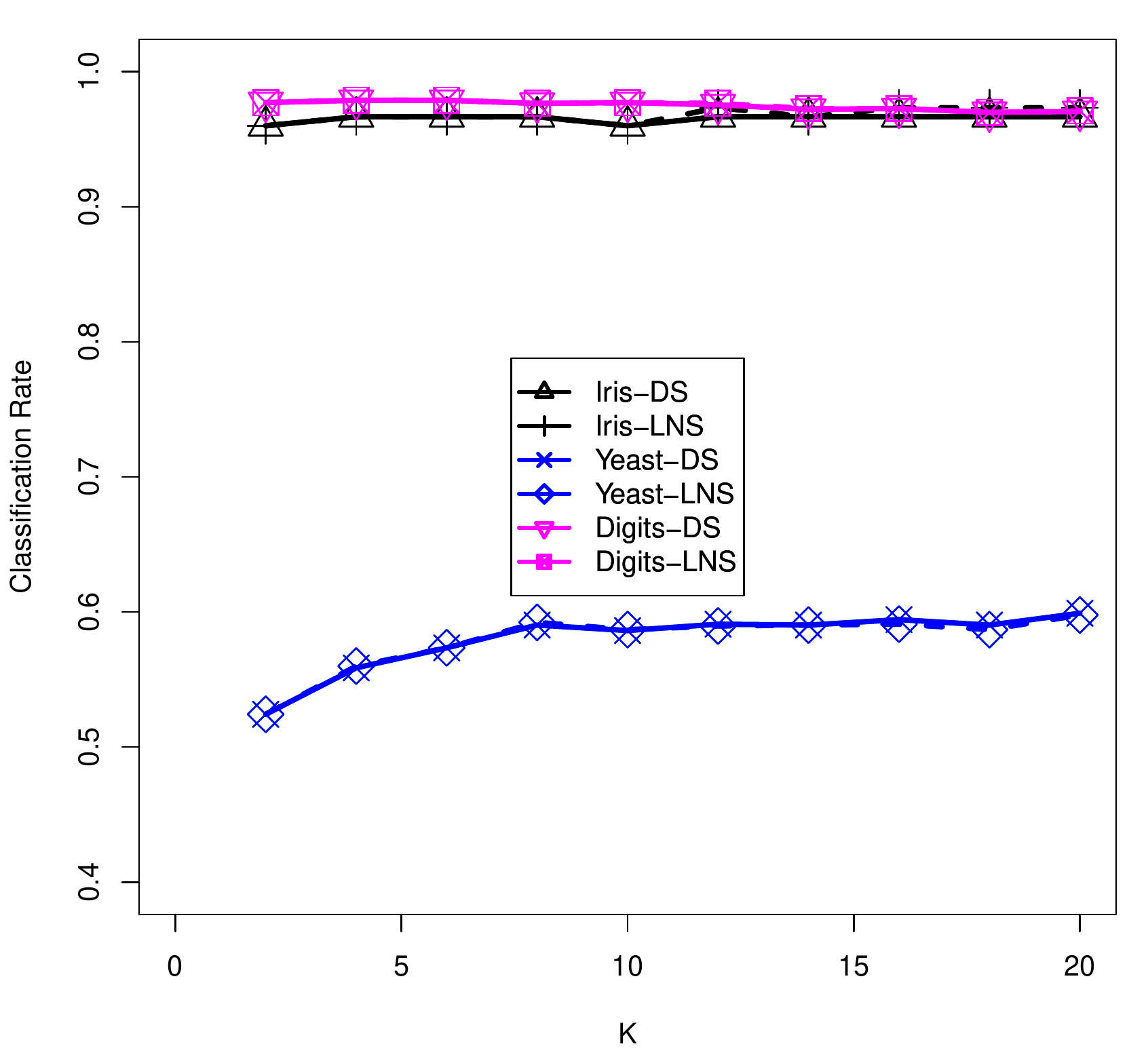}\hfill
\caption{Classification results with different values of $K$
on UCI data set. In the figure, the legend ``Iris-DS" means it is the classification rates
on Iris data set using DS combination rule. Same as the other legends.} \label{Exp5classificationrateUCI} \end{figure} \end{center}
\begin{center} \begin{figure}[!thbt] \centering
       \includegraphics[width=0.8\linewidth]{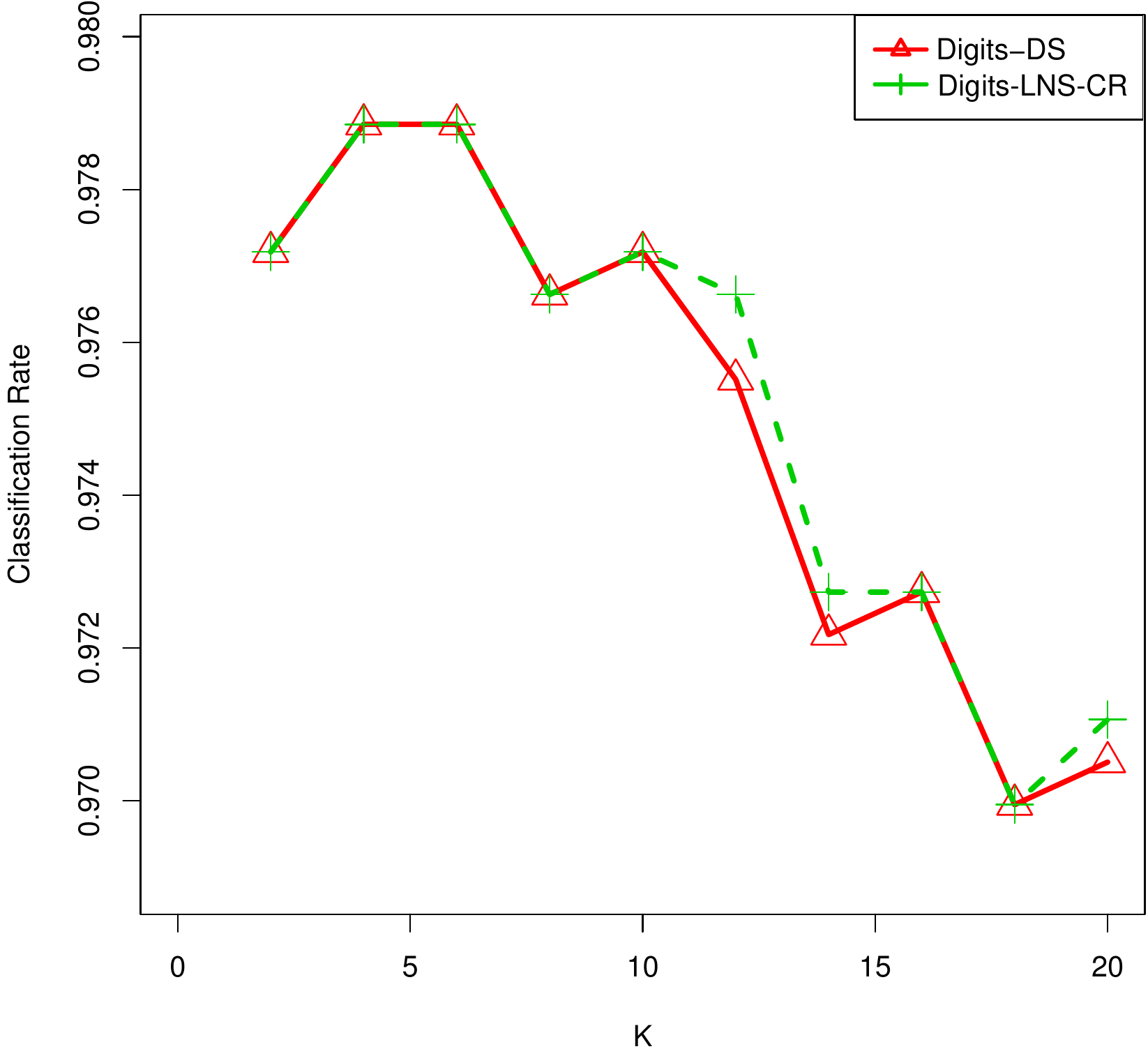} \hfill
\caption{Classification rates on Digits data set.} \label{Exp5classificationrateDigits} \end{figure} \end{center}
\begin{center} \begin{figure}[!thbt] \centering
		\includegraphics[width=0.8\linewidth]{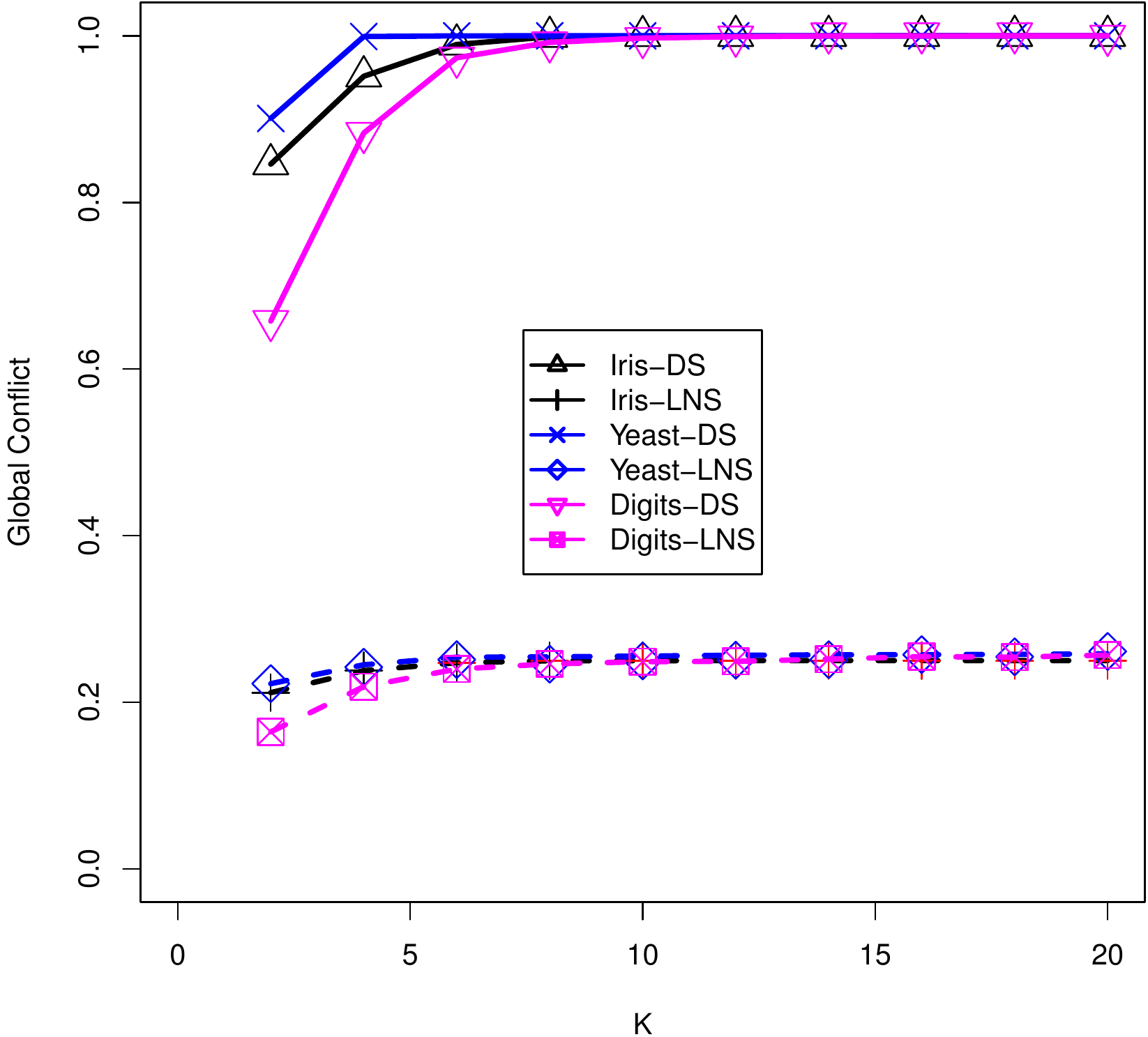}\hfill
\caption{Global conflict using conjunctive rule and $\LNS$ rule varying with
different values of $K$. In the figure, the legend ``Iris-DS" means it is the
conflict on Iris data set using DS combination rule. Same as the other legends.} \label{digits_conflict} \end{figure} \end{center}
}

{
\subsection{Perspective}
The above two examples are just two perspectives on the application of $\LNS$ rule. In the first example,
there are some special noisy data in the training data set. At this time, the sources should not be
considered with equal reliability. In this situation, using the DS rule or the conjunctive rule in EKNN model could not
get good results. In the second example, \change{it is shown that the global conflict may
tend to one quickly  as $K$ increases.} Sometimes we even could not do the normalization process for DS rule because of
the machine precision.
}

In  real world social networks, \change{the available information can be  uncertain, or even
noisy.} At this time, if we want to do a classification task such as for recommendation, the conjunctive rule could not be
applied as the sources are not all reliable. Even if the sources are reliable, the global conflict may tend to 1 quickly if the bbas are
not consistent. 
At this time, $\LNS$ rule can be an alternative choice. In the future work, we will
study how Dempster's degree of conflict is distributed in the feature space, and to study what special
information contained in the moderate  degree of global conflict kept by  $\LNS$ rule.

\section{Conclusion}
Uncertainty in big data applications has attracted more and more {attention}.
The theory of belief functions is one of the uncertainty theories
{allowing a model to} deal with imprecise and uncertain information.
This theory is also well designed for information fusion. However, despite that a lot of
combination rules have been proposed in  recent years in this framework, they are not able to combine
a large number of {sources} because of the complexity or the absorbing element.

In this paper, a new combination rule, {named} $\LNS$ rule, preserving the principle of the conjunctive rule is proposed. This rule considers the mass functions given by the sources and groups them according to their set of focal elements
(without auto-conflict). The mass functions of each group can be summarized by one mass function after combination. The reliability of the source is estimated by the proportion of bbas in one group. Therefore, after discounting the mass function of each group by the reliability factor, the final combination can be proceeded  by the conjunctive rule (or another rule according to the application). If the number of sources in each group is high enough, an approximation method is presented.

The $\LNS$ rule  is able to combine a large number of sources. The only existing method allowing to combine a large number of mass functions is the
average rule. However, that rule may give more importance to few sources with a high belief (even if the source is not reliable) and
cannot capture the conflict between the sources. The proposed rule with a reasonable complexity (lower than the $\DP$ and $\PCR$ rules)
can provide good combination results.

%

Overall, this work provides a perspective for the application of belief functions on big data. We will study how to apply  $\LNS$ rule
on the problems of social network and crowdsourcing in the future research work.

\section*{Acknowledgements}
This work was supported by the National Natural Science Foundation of China (Nos.61701409, 61135001, 61403310, 61672431), the Natural
Science Basic Research Plan in Shaanxi Province of China (No.2018JQ6005), and the Fundamental Research Funds for the Central Universities of China (No.3102016QD088).

\bibliographystyle{IEEEtranN}
\bibliography{paperlist}

\begin{thebibliography}{54}
\providecommand{\natexlab}[1]{#1}
\providecommand{\url}[1]{#1}
\csname url@samestyle\endcsname
\providecommand{\newblock}{\relax}
\providecommand{\bibinfo}[2]{#2}
\providecommand{\BIBentrySTDinterwordspacing}{\spaceskip=0pt\relax}
\providecommand{\BIBentryALTinterwordstretchfactor}{4}
\providecommand{\BIBentryALTinterwordspacing}{\spaceskip=\fontdimen2\font plus
\BIBentryALTinterwordstretchfactor\fontdimen3\font minus
  \fontdimen4\font\relax}
\providecommand{\BIBforeignlanguage}[2]{{%
\expandafter\ifx\csname l@#1\endcsname\relax
\typeout{** WARNING: IEEEtranN.bst: No hyphenation pattern has been}%
\typeout{** loaded for the language `#1'. Using the pattern for}%
\typeout{** the default language instead.}%
\else
\language=\csname l@#1\endcsname
\fi
#2}}
\providecommand{\BIBdecl}{\relax}
\BIBdecl

\bibitem[Zhou et~al.(2017)Zhou, Martin, and Pan]{zhou2017evidence}
K.~Zhou, A.~Martin, and Q.~Pan, ``Evidence combination for a large number of
  sources,'' in \emph{20th International Conference on Information
  Fusion}.\hskip 1em plus 0.5em minus 0.4em\relax IEEE, 2017, pp. 1--8.

\bibitem[Den{\oe}ux(1995)]{denoeux1995k}
T.~Den{\oe}ux, ``A $k$--nearest neighbor classification rule based on
  dempster-shafer theory,'' \emph{Systems, Man and Cybernetics, IEEE
  Transactions on}, vol.~25, no.~5, pp. 804--813, 1995.

\bibitem[Deng et~al.(2016)Deng, Liu, Deng, and Mahadevan]{deng2016improved}
X.~Deng, Q.~Liu, Y.~Deng, and S.~Mahadevan, ``An improved method to construct
  basic probability assignment based on the confusion matrix for classification
  problem,'' \emph{Information Sciences}, vol. 340, pp. 250--261, 2016.

\bibitem[Masson and Den{\oe}ux(2008)]{masson2008ecm}
M.-H. Masson and T.~Den{\oe}ux, ``{ECM}: An evidential version of the fuzzy
  $c$-means algorithm,'' \emph{Pattern Recognition}, vol.~41, no.~4, pp.
  1384--1397, 2008.

\bibitem[Zhou et~al.(2016)Zhou, Martin, Pan, and Liu]{zhou2016ecmdd}
K.~Zhou, A.~Martin, Q.~Pan, and Z.-G. Liu, ``Ecmdd: Evidential $c$-medoids
  clustering with multiple prototypes,'' \emph{Pattern Recognition}, vol.~60,
  pp. 239 -- 257, 2016.

\bibitem[Zhou et~al.(2015)Zhou, Martin, Pan, and Liu]{zhou2015median}
K.~Zhou, A.~Martin, Q.~Pan, and Z.-g. Liu, ``Median evidential $c$-means
  algorithm and its application to community detection,'' \emph{Knowledge-Based
  Systems}, vol.~74, pp. 69--88, 2015.

\bibitem[Smets(2007)]{Smets07a}
P.~Smets, ``Analyzing the combination of conflicting belief functions,''
  \emph{Information Fusion}, vol.~8, pp. 387--412, 2007.

\bibitem[Smets(1990)]{smets1990combination}
------, ``The combination of evidence in the transferable belief model,''
  \emph{Pattern Analysis and Machine Intelligence, IEEE Transactions on},
  vol.~12, no.~5, pp. 447--458, 1990.

\bibitem[Smets and Kennes(1994)]{smets1994transferable}
P.~Smets and R.~Kennes, ``The transferable belief model,'' \emph{Artificial
  intelligence}, vol.~66, no.~2, pp. 191--234, 1994.

\bibitem[Martin et~al.(2008{\natexlab{a}})Martin, Jousselme, and
  Osswald]{martin2008conflict}
A.~Martin, A.-L. Jousselme, and C.~Osswald, ``Conflict measure for the
  discounting operation on belief functions,'' in \emph{Information Fusion,
  2008 11th International Conference on}.\hskip 1em plus 0.5em minus
  0.4em\relax IEEE, 2008, pp. 1--8.

\bibitem[Liu(2006)]{liu2006analyzing}
W.~Liu, ``Analyzing the degree of conflict among belief functions,''
  \emph{Artificial Intelligence}, vol. 170, no.~11, pp. 909--924, 2006.

\bibitem[Destercke and Burger(2013)]{destercke2013toward}
S.~Destercke and T.~Burger, ``Toward an axiomatic definition of conflict
  between belief functions,'' \emph{Cybernetics, IEEE Transactions on},
  vol.~43, no.~2, pp. 585--596, 2013.

\bibitem[Martin and Osswald(2006)]{Martin06a}
A.~Martin and C.~Osswald, ``Human experts fusion for image classification,''
  \emph{Information \& Security: An International Journal, Special issue on
  Fusing Uncertain, Imprecise and Paradoxist Information (DSmT)}, vol.~20, pp.
  122--143, 2006.

\bibitem[Lef{\`e}vre and Elouedi(2013)]{lefevre2013preserve}
E.~Lef{\`e}vre and Z.~Elouedi, ``How to preserve the conflict as an alarm in
  the combination of belief functions,'' \emph{Decision Support Systems},
  vol.~56, pp. 326--333, 2013.

\bibitem[Yager(1987)]{yager1987dempster}
R.~R. Yager, ``On the dempster-shafer framework and new combination rules,''
  \emph{Information sciences}, vol.~41, no.~2, pp. 93--137, 1987.

\bibitem[Dubois and Prade(1988)]{dubois1988representation}
D.~Dubois and H.~Prade, ``Representation and combination of uncertainty with
  belief functions and possibility measures,'' \emph{Computational
  Intelligence}, vol.~4, no.~3, pp. 244--264, 1988.

\bibitem[Ilin and Blasch(2015)]{ilin2015information}
R.~Ilin and E.~Blasch, ``\change{Information fusion with belief functions: A
  comparison of proportional conflict redistribution PCR5 and PCR6 rules for
  networked sensors},'' in \emph{\change{18th International Conference on
  Information Fusion}}.\hskip 1em plus 0.5em minus 0.4em\relax IEEE, 2015, pp.
  2084--2091.

\bibitem[Martin et~al.(2008{\natexlab{b}})Martin, Osswald, Dezert, and
  Smarandache]{martin2008general}
A.~Martin, C.~Osswald, J.~Dezert, and F.~Smarandache, ``\change{General
  combination rules for qualitative and quantitative beliefs},''
  \emph{\change{Journal of Advances in Information Fusion}}, vol.~3, no.~2, pp.
  67--89, 2008.

\bibitem[Zhao et~al.(2016)Zhao, Jia, and Shi]{zhao2016novel}
Y.~Zhao, R.~Jia, and P.~Shi, ``A novel combination method for conflicting
  evidence based on inconsistent measurements,'' \emph{Information Sciences},
  vol. 367--368, pp. 125--142, 2016.

\bibitem[Orponen(1990)]{Orponen90a}
P.~Orponen, ``Dempster's rule of combination is \#{$P$}-complete,''
  \emph{Artificial Intelligence}, vol.~44, pp. 245--253, 1990.

\bibitem[Da~Silva and Milidi{\'u}(1992)]{daSilva92a}
W.~T. Da~Silva and R.~L. Milidi{\'u}, ``Algorithms for combining belief
  functions,'' \emph{International Journal of Approximate Reasoning}, vol.~7,
  no. 1-2, pp. 73 -- 94, 1992.

\bibitem[Martin and Osswald(2007)]{martin2007toward}
A.~Martin and C.~Osswald, ``Toward a combination rule to deal with partial
  conflict and specificity in belief functions theory,'' in \emph{10th
  International Conference on Information Fusion}.\hskip 1em plus 0.5em minus
  0.4em\relax IEEE, 2007, pp. 1--8.

\bibitem[Smets(1997)]{smets97alpha}
P.~Smets, ``The $\alpha$-junctions: the commutative and associative non
  interactive combination operators applicable to belief function,'' in
  \emph{1st International Joint Conference on Qualitative and Quantitative
  Practical Reasoning}, 1997, pp. 131--153.

\bibitem[Leung et~al.(2013)Leung, Ji, and Ma]{leung2013integrated}
Y.~Leung, N.-N. Ji, and J.-H. Ma, ``An integrated information fusion approach
  based on the theory of evidence and group decision-making,''
  \emph{Information Fusion}, vol.~14, no.~4, pp. 410--422, 2013.

\bibitem[Shafer(1976)]{ds2}
G.~Shafer, \emph{A mathematical theory of evidence}.\hskip 1em plus 0.5em minus
  0.4em\relax Princeton University Press, 1976.

\bibitem[Smets(2002)]{smets2002application}
P.~Smets, ``The application of the matrix calculus to belief functions,''
  \emph{International Journal of Approximate Reasoning}, vol.~31, no.~1, pp.
  1--30, 2002.

\bibitem[Smets(2005)]{smets2005decision}
------, ``Decision making in the {TBM}: the necessity of the pignistic
  transformation,'' \emph{International Journal of Approximate Reasoning},
  vol.~38, no.~2, pp. 133--147, 2005.

\bibitem[Sentz and Ferson(2002)]{sentz2002combination}
K.~Sentz and S.~Ferson, ``Combination of evidence in dempster-shafer theory,''
  SAndia National Laboratorie, Tech. Rep., 2002.

\bibitem[Smets(1993{\natexlab{a}})]{Smets93a}
P.~Smets, ``{B}elief {F}unctions: the {D}isjunctive {R}ule of {C}ombination and
  the {G}eneralized {B}ayesian {T}heorem,'' \emph{International Journal of
  Approximate Reasoning}, vol.~9, pp. 1--35, 1993.

\bibitem[Martin(2005)]{martin2005comparative}
A.~Martin, ``Comparative study of information fusion methods for sonar images
  classification,'' in \emph{Information Fusion, 2005 8th International
  Conference on}, vol.~2.\hskip 1em plus 0.5em minus 0.4em\relax IEEE, 2005,
  pp. 7--pp.

\bibitem[Samet et~al.(2013)Samet, Lefevre, and Ben~Yahia]{samet2013reliability}
A.~Samet, E.~Lefevre, and S.~Ben~Yahia, ``Reliability estimation with extrinsic
  and intrinsic measure in belief function theory,'' in \emph{5th International
  Conference on Modeling, Simulation and Applied Optimization}.\hskip 1em plus
  0.5em minus 0.4em\relax IEEE, 2013, pp. 1--6.

\bibitem[Schubert(2011)]{schubert2011conflict1}
J.~Schubert, ``Conflict management in dempster--shafer theory using the degree
  of falsity,'' \emph{International Journal of Approximate Reasoning}, vol.~52,
  no.~3, pp. 449--460, 2011.

\bibitem[Elouedi et~al.(2001)Elouedi, Mellouli, and
  Smets]{elouedi2001evaluation}
Z.~Elouedi, K.~Mellouli, and P.~Smets, ``The evaluation of sensors' reliability
  and their tuning for multisensor data fusion within the transferable belief
  model,'' in \emph{Symbolic and Quantitative Approaches to Reasoning with
  Uncertainty}.\hskip 1em plus 0.5em minus 0.4em\relax Springer, 2001, pp.
  350--361.

\bibitem[Samet et~al.(2015)Samet, Lef{\`e}vre, Hammami, and
  Ben~Yahia]{samet2015reliability}
A.~Samet, E.~Lef{\`e}vre, I.~Hammami, and S.~Ben~Yahia, ``Reliability
  estimation measure: Generic discounting approach,'' \emph{International
  Journal of Pattern Recognition and Artificial Intelligence}, vol.~29, no.~07,
  p. 1559011, 2015.

\bibitem[Yang et~al.(2013)Yang, Han, and Han]{yang2013discounted}
Y.~Yang, D.~Han, and C.~Han, ``Discounted combination of unreliable evidence
  using degree of disagreement,'' \emph{International Journal of Approximate
  Reasoning}, vol.~54, no.~8, pp. 1197--1216, 2013.

\bibitem[Klein and Colot(2011)]{klein2011singular}
J.~Klein and O.~Colot, ``Singular sources mining using evidential conflict
  analysis,'' \emph{International Journal of Approximate Reasoning}, vol.~52,
  no.~9, pp. 1433--1451, 2011.

\bibitem[Smets(1995)]{smets1995canonical}
P.~Smets, ``The canonical decomposition of a weighted belief,'' in \emph{14th
  International Joint Conference on Artificial Intelligence}, vol.~95, 1995,
  pp. 1896--1901.

\bibitem[Den{\oe}ux(2008)]{denoeux2008conjunctive}
T.~Den{\oe}ux, ``Conjunctive and disjunctive combination of belief functions
  induced by nondistinct bodies of evidence,'' \emph{Artificial Intelligence},
  vol. 172, no.~2, pp. 234--264, 2008.

\bibitem[Ke et~al.(2014)Ke, Ma, and Wang]{Ke14a}
X.~Ke, L.~Ma, and Y.~Wang, ``Some notes on canonical decomposition and
  separability of a belief function,'' in \emph{Belief Functions: Theory and
  Applications}, ser. Lecture Notes in Computer Science, F.~Cuzzolin, Ed., vol.
  8764.\hskip 1em plus 0.5em minus 0.4em\relax Springer International
  Publishing, 2014, pp. 153--160.

\bibitem[Kennes(1992)]{kennes1992computational}
R.~Kennes, ``Computational aspects of the m{\"o}bius transformation of
  graphs,'' \emph{Systems, Man and Cybernetics, IEEE Transactions on}, vol.~22,
  no.~2, pp. 201--223, 1992.

\bibitem[Dempster(1967)]{dempster1967upper}
A.~P. Dempster, ``Upper and lower probabilities induced by a multivalued
  mapping,'' \emph{The annals of mathematical statistics}, pp. 325--339, 1967.

\bibitem[Osswald and Martin(2006)]{osswald2006understanding}
C.~Osswald and A.~Martin, ``Understanding the large family of dempster-shafer
  theory's fusion operators-a decision-based measure,'' in \emph{9th
  International Conference on Information Fusion}.\hskip 1em plus 0.5em minus
  0.4em\relax IEEE, 2006, pp. 1--7.

\bibitem[Smets(1993{\natexlab{b}})]{smets1993belief}
P.~Smets, ``Belief functions: the disjunctive rule of combination and the
  generalized bayesian theorem,'' \emph{International Journal of approximate
  reasoning}, vol.~9, no.~1, pp. 1--35, 1993.

\bibitem[Wilson(2000)]{Wilson00a}
N.~Wilson, ``Algorithms for {D}empster-{S}hafer theory,'' in \emph{Hanbook of
  defeasible reasoning and uncertainty management}, D.~Gabbay and P.~Smets,
  Eds.\hskip 1em plus 0.5em minus 0.4em\relax Boston: Kluwer Academic
  Publisher, 2000, vol. 5: Algorithms for uncertainty and Defeasible Reasoning,
  pp. 421--475.

\bibitem[Den{\oe}ux and Ben~Yaghlane(2002)]{Denoeux02a}
T.~Den{\oe}ux and A.~Ben~Yaghlane, ``Approximating the combination of belief
  functions using the fast {M\"o}bius transform in a coarsened frame,''
  \emph{International Journal of Approximate Reasoning}, vol.~30, no. 1-2, pp.
  77--101, 2002.

\bibitem[Martin(2009)]{Martin09a}
A.~Martin, ``Implementing general belief function framework with a practical
  codification for low complexity,'' in \emph{Advances and Applications of DSmT
  for Information Fusion}, F.~Smarandache and J.~Dezert, Eds.\hskip 1em plus
  0.5em minus 0.4em\relax American Research Press Rehoboth, 2009, vol.~3,
  ch.~7, pp. 217--274.

\bibitem[Den{\oe}ux(2006)]{denoeux2006cautious}
T.~Den{\oe}ux, ``The cautious rule of combination for belief functions and some
  extensions,'' in \emph{9th International Conference on Information
  Fusion}.\hskip 1em plus 0.5em minus 0.4em\relax IEEE, 2006, pp. 1--8.

\bibitem[Chin and Fu(2015)]{chin2015weighted}
K.-S. Chin and C.~Fu, ``Weighted cautious conjunctive rule for belief functions
  combination,'' \emph{Information Sciences}, vol. 325, pp. 70--86, 2015.

\bibitem[Murphy(2000)]{murphy2000combining}
C.~K. Murphy, ``Combining belief functions when evidence conflicts,''
  \emph{Decision support systems}, vol.~29, no.~1, pp. 1--9, 2000.

\bibitem[Smarandache and Dezert(2004--2009)]{smarandache2006advances}
F.~Smarandache and J.~Dezert, \emph{Advances and Applications of DSmT for
  Information Fusion}.\hskip 1em plus 0.5em minus 0.4em\relax American Research
  Press, Rehoboth, 2004--2009, vol. 1--3.

\bibitem[Dubois and Prade(1990)]{Dubois90a}
D.~Dubois and H.~Prade, ``Consonant approximation of belief functions,''
  \emph{International Journal of Approximate Reasoning}, vol.~4, no. 5-6, pp.
  419--449, 1990.

\bibitem[Aregui and Den{\oe}ux(2008)]{Aregui08a}
A.~Aregui and T.~Den{\oe}ux, ``Constructing consonant belief functions from
  sample data using confidence sets of pignistic probabilities,''
  \emph{International Journal of Approximate Reasoning}, vol.~49, no.~3, pp.
  575 -- 594, 2008.

\bibitem[Zhou and Martin(2015)]{ibelief}
\BIBentryALTinterwordspacing
K.~Zhou and A.~Martin, \emph{ibelief: Belief Function Implementation}, 2015, r
  package version 1.2. [Online]. Available:
  \url{http://CRAN.R-project.org/package=ibelief}
\BIBentrySTDinterwordspacing

\bibitem[Burger and Destercke(2013)]{burger2013randomly}
T.~Burger and S.~Destercke, ``How to randomly generate mass functions,''
  \emph{International Journal of Uncertainty, Fuzziness and Knowledge-Based
  Systems}, vol.~21, no.~05, pp. 645--673, 2013.

\end{thebibliography}
\end{document}